\documentclass[review]{elsarticle}
\usepackage{amsmath,amsfonts}
\usepackage{algorithmic}
\usepackage{array}
\usepackage[caption=false,font=normalsize,labelfont=sf,textfont=sf]{subfig}
\usepackage{textcomp}
\usepackage{stfloats}
\usepackage{hyperref}
\usepackage{pgfplots}
\usepackage{filecontents}
\usepackage{latexsym} 
\usepackage{graphics}
\usepackage{epsfig}
\usepackage{graphicx}
\usepackage{tikz-qtree}
\usepackage{tikz}
\usepackage{amsmath}
\usepackage{dirtytalk} 
\usepackage{bbm} 
\usepackage{changepage}
\usepackage{pgfplots}
\usepgfplotslibrary{groupplots}
\usepackage[utf8]{inputenc}
\usepackage[english]{babel}
\usepackage{rotating}
\usepackage{booktabs}
\usepackage{cellspace}
\usepgfplotslibrary{statistics}
\pgfmathsetmacro{\offset}{0.05}
\pgfplotsset{width=11cm,compat=newest}

\usepackage{tikz}
\usetikzlibrary{arrows.meta, positioning}

\usepackage{varwidth}
\usepackage{booktabs} 
\usepackage{epsfig}
\usepackage{amsfonts}
\usepackage{array}

\usepackage{rotating}
\usepackage{tabularx}
\usepackage{color}
\usepackage{multirow}
\usepackage{url}
\usepackage{lscape}
\usepackage{subcaption}
\usepackage{xargs} 
\newcolumntype{x}[1]{>{\centering\let\newline\\\arraybackslash\hspace{0pt}}p{#1}}

\usepackage{tabularx}
\usepackage{array}
\usepackage{colortbl}
\usepackage{multirow}
\usepackage{datetime}
\usepackage[mathscr]{eucal}
\usepackage{amssymb}
\usepackage{amsmath}
\usepackage{derivative}
\usepackage[utf8]{inputenc}
\usepackage[T1]{fontenc}
\usepackage{mathtools}
\usepackage[thinc]{esdiff}
\usepackage{wrapfig}
\usepackage{caption}
\usepackage{lscape}
\usepackage{rotating}
\usepackage{url}
\usepackage{babel}
\usepackage{translator}
\usepackage{pgfkeys,pgfcalendar}
\usepackage{multirow}
\usepackage{hhline}
\usepackage{diagbox}
\usepackage{eurosym}
\usepackage{amssymb}
\usepackage{pifont}
\graphicspath{ {/home/ciaran/Downloads/} }
\usepackage{acronym}
\usepackage{booktabs}
\usepackage{url}
\usepackage{verbatim}
\usepackage{graphicx}

\usepackage{hyperref}

\usepackage{latexsym} 
\usepackage{graphics}
\usepackage{epsfig}
\usepackage{graphicx}
\usepackage{tikz-qtree}
\usepackage{tikz}
\usepackage{amsmath}
\usepackage{dirtytalk} 
\usepackage{bbm} 
\usepackage{pgfplots}
\usepackage{caption}
\usepgfplotslibrary{groupplots}
\usepackage[utf8]{inputenc}
\usepackage[english]{babel}
\usepackage{pgfplots}
\pgfplotsset{compat=1.17}
\usepackage{rotating}
\pgfplotsset{width=11cm,compat=newest}
\makeatletter
\def\BState{\State\hskip-\ALG@thistlm}
\makeatother
\usepackage{graphicx}
\usepackage{subfig}
\usepackage{tikz}
\usepackage{pgfplots}

\usepackage{varwidth}
\usepackage{booktabs} 
\usepackage{epsfig}
\usepackage{amsfonts}
\usepackage{array}

\usepackage{rotating}
\usepackage{tabularx}
\usepackage{color}
\usepackage{multirow}
\usepackage{url}
\usepackage{lscape}
\usepackage{xargs} 
\newcolumntype{x}[1]{>{\centering\let\newline\\\arraybackslash\hspace{0pt}}p{#1}}

\usepackage{tabularx}
\usepackage{array}
\usepackage{colortbl}
\usepackage{multirow}
\usepackage{datetime}
\usepackage[mathscr]{eucal}
\usepackage{amssymb}
\usepackage{amsmath}
\usepackage{graphicx}
\usepackage{derivative}
\usepackage[utf8]{inputenc}
\usepackage[T1]{fontenc}
\usepackage{mathtools}
\usepackage[thinc]{esdiff}
\usepackage{wrapfig}
\usepackage{caption}
\usepackage{lscape}
\usepackage{rotating}
\usepackage{url}
\usepackage{babel}
\usepackage{translator}
\usepackage{pgfkeys,pgfcalendar}
\usepackage{graphicx}
\usepackage{multirow}
\usepackage{hhline}
\usepackage{diagbox}
\usepackage{eurosym}
\usepackage{amssymb}
\usepackage{pifont}
\graphicspath{ {/home/ciaran/Downloads/} }
\usepackage{acronym}
\usepackage{booktabs}
\def\preparecolorrefs#1{%
  \setcounter{refindex}{0}%
  \whiledo{\value{refindex}<#1}{%
    \stepcounter{refindex}%
    \expandafter\def\csname\therefindex color\endcsname{black}%
  }%
}
\usepackage{hyperref}

\journal{a Journal}

\biboptions{authoryear}

\begin{document}

\begin{frontmatter}

\title{Conformal Prediction for Electricity Price Forecasting in the Day-Ahead and Real-Time Balancing Market}

\author{Ciaran O'Connor}
\address{SFI CRT in Artificial Intelligence, School of Computer Science \& IT, University College Cork, Ireland}
\ead{119226305@umail.ucc.ie}

\author{Mohamed Bahloul}
\address{International Energy Research Centre, Tyndall National Institute, Cork, Ireland}
\ead{mohamed.bahloul@ierc.ie}

\author{Roberto Rossi}
\address{Business School, University of Edinburgh, Edinburgh, UK}
\ead{roberto.rossi@ed.ac.uk}

\author{Steven Prestwich, Andrea Visentin}
\address{SFI Insight Centre for Data Analytics, School of Computer Science \& IT, University College Cork, Ireland}
\ead{s.prestwich@cs.ucc.ie, andrea.visentin@ucc.ie}

\begin{abstract}
The integration of renewable energy into electricity markets poses significant challenges to price stability and increases the complexity of market operations. Accurate and reliable electricity price forecasting is crucial for effective market participation, where price dynamics can be significantly more challenging to predict. Probabilistic forecasting, through prediction intervals, efficiently quantifies the inherent uncertainties in electricity prices, supporting better decision-making for market participants. 
This study explores the enhancement of probabilistic price prediction using Conformal Prediction (CP) techniques, specifically Ensemble Batch Prediction Intervals and Sequential Predictive Conformal Inference. These methods provide precise and reliable prediction intervals, outperforming traditional models in validity metrics.  We propose an ensemble approach that combines the efficiency of quantile regression models with the robust coverage properties of time series adapted CP techniques. This ensemble delivers both narrow prediction intervals and high coverage, leading to more reliable and accurate forecasts. 
We further evaluate the practical implications of CP techniques through a simulated trading algorithm applied to a battery storage system. The ensemble approach demonstrates improved financial returns in energy trading in both the Day-Ahead and Balancing Markets, highlighting its practical benefits for market participants.
\end{abstract}


\end{frontmatter}

\section{Introduction}
Electricity Price Forecasting (EPF) is particularly challenging due to market volatility, sudden price fluctuations, and seasonal demand shifts. Accurate EPF is essential for all market participants in deregulated markets, providing a foundation for informed decision-making and financial planning. Power generation companies rely on forecasts to hedge against risks from deviations between day-ahead and real-time markets, thereby stabilising revenue streams and operational planning. Load-serving utilities use forecasts to secure energy at lower prices, optimising procurement strategies and ensuring cost-effective supply for consumers. Virtual traders, lacking actual generation power or load requirements, depend on EPF to estimate price spreads between spot and forward markets, exploiting arbitrage opportunities to generate profits. The increasing integration of renewable energy sources and the advent of smart grids further complicate EPF, exacerbating market volatility.

Spot markets such as the Day-Ahead Market (DAM), Intra-Day Market (IDM), and Balancing Market (BM) are crucial for maintaining grid stability. Yet, increasing market price volatility, driven by forecast errors in renewable electricity generation, presents significant challenges \cite{ortner2019future}. This underscores the critical importance of accurate forecasting in stabilising energy production planning, optimising trading strategies, and implementing risk-aware policies to ensure a reliable, cost-effective, and sustainable energy supply. Accurate EPF supports market participants' financial health and plays an important role in advancing energy policy goals surrounding renewable integration, grid stability, and reducing carbon emissions.

There are two major EPF schemes: point forecasts and probabilistic forecasts. Point forecasts provide single price predictions that are easy to interpret and establish parsimonious predictor-target relationships under the assumption of homoscedasticity. However, they cannot quantify uncertainties, as they typically link predictors to target values using the conditional mean function. Thus, a critical evolution in EPF has been the transition from traditional point forecasts to probabilistic forecasts which offer a range of potential outcomes, quantifying uncertainty, and generating prediction intervals (PIs) that reflect the confidence levels of forecasts. Probabilistic Electricity Price Forecasting (PEPF) can capture the full range of potential future prices in dynamic, non-linear markets \cite{nowotarski2018recent, ziel2018probabilistic} addressing uncertainties in smart grids, supply-demand dynamics, and price variations emphasising the operational impact of forecasts \cite{khajeh2022applications}.

Key methods in probabilistic forecasting include Quantile Regression (QR), Quantile Regression Averaging (QRA), Bayesian, Historical Simulation, and Bootstrapped PIs. However, these methods fail to consistently provide validity, particularly in non-stationary settings, which limits their reliability in high-stakes applications. Conformal Prediction (CP) has gained significant attention for its ability to provide valid and adaptive PIs by ensuring predefined confidence levels are met, regardless of the underlying data distribution or non-stationarity  \cite{gammerman1998learning, vovk2005algorithmic}. Recent advancements, such as Ensemble Batch Prediction Intervals (EnbPI) \cite{xu2021conformal} and Sequential Predictive Conformal Inference (SPCI) \cite{xu2023sequential}, have successfully extended CP methodologies to time series applications, addressing the unique challenges of electricity markets. This focus on validity, combined with adaptability, positions CP as a robust framework for managing uncertainty in highly volatile and dynamic markets, making it a critical tool for risk management.


\subsection{Contributions}
Building on the existing literature in the field, this paper makes several notable contributions to the area of PEPF in the DAM and BM. While the IDM and BM share structural similarities, we focus primarily on the BM due to its fundamental role in balancing real-time supply and demand. The key contributions of our research are outlined as follows:
\begin{itemize}
    \item \textit{Application of QRA and CP to BM}: While QRA and Split Conformal Prediction (SCP) have been applied to the DAM, their use in the BM remains unexplored. Our study address this gap by applying these methods to both markets, exploring their performance and how combining QRA with CP impacts model accuracy and reliability.
    
    \item \textit{Introduction of Time Series Adapted CP Approaches}: We employ EnbPI and SPCI in both DAM and BM contexts. To our knowledge, these CP approaches have not been previously applied to any PEPF context. Our research demonstrates the potential benefits of these methods in providing reliable PIs.
    
    \item \textit{Introduction of Q-Ens}: We propose a novel ensemble method that combines predictions from QR, EnbPI, and SPCI models. This approach aims to balance the superior efficiency of QR with the superior validity of CP methods, offering a robust solution in PEPF. 
    
    \item \textit{Evaluation and Trading Impact}: We assess the efficiency and validity of forecasts using various probabilistic metrics, and investigate the importance of coverage guarantees for trading outcomes in both DAM and BM. Our financial analysis includes both single-trade and high-frequency trading strategies, examining the impact of probabilistic forecasts on trading decisions and outcomes.
\end{itemize}
This research advances the field by introducing innovative CP and QRA applications, proposing a new ensemble approach, and comprehensively evaluating probabilistic forecasts in electricity markets. We provide the accompanying dataset and Python code for forecasting and trading, facilitating the adoption and replication of our methods. Our findings highlight the strengths of CP methods in ensuring coverage guarantees and benefiting trading strategies.

The structure of this paper is as follows: Section \ref{pepfliteraturereview} provides an overview of recent advancements in PEPF within the context of the DAM and BM. Section \ref{Meth}, we detail our methodological framework, including our forecasting models and probabilistic approaches. Section \ref{EvalMetrics} discusses probabilistic metrics used to evaluate the efficiency and validity of our forecasts. The effectiveness of each forecasting approach, especially the proposed ensemble, is discussed through case study results in Section \ref{resultssec}. Finally, Section \ref{conclusionsec} summarises our findings.


\section{Related Work}\label{pepfliteraturereview}
EPF has emerged as a critical research area due to the increasing complexity and volatility of electricity markets. Accurate forecasting is essential for market participants to make informed decisions, mitigate financial risks, and optimise operations. Probabilistic forecasting methods, in particular, address the inherent uncertainties in these markets by providing PIs rather than single-point estimates, allowing stakeholders to assess the range of potential outcomes. Table \ref{TableLR} provides a comprehensive summary of the literature on probabilistic EPF methods, highlighting the evolution from traditional point forecasts to advanced probabilistic techniques, including QR and CP. This section critically examines the role of forecasting in electricity spot markets, foundational concepts of uncertainty in EPF, recent advancements in probabilistic methods, and the growing role of CP as an alternative framework, while evaluating their strengths, limitations, and practical implications.

\subsection{Electricity Spot Markets}
The DAM is a cornerstone of electricity trading, where market participants submit bids for electricity trades up to one day in advance, with the bidding process closing in the afternoon to determine market-clearing prices and volumes based on supply, demand, and grid constraints \cite{newbery2016benefits, ilea2018european}. 
While the DAM enhances cross-border electricity trading and grid stability, it faces challenges from forecast uncertainties driven by the increasing penetration of intermittent renewable energy sources such as wind and solar \cite{miraki2025probabilistic}. These challenges necessitate advanced forecasting models capable of managing the variability introduced by weather-dependent renewables, such as privacy-preserving frameworks that enhance accuracy by integrating diverse data sources \cite{liu2025probabilistic}. Initially dominated by traditional statistical methods, forecasting has shifted toward advanced machine learning (ML) techniques, which excel at capturing the complex, non-linear relationships inherent in electricity markets \cite{ugurlu2018electricity, lago2018forecasting, chen2019brim, li2021day}. Hybrid approaches combining ML with other techniques have further improved precision, particularly in modelling supply-demand dynamics and adapting to varying market conditions \cite{lago2021forecasting, elrobrini2024federated}.

The IDM is crucial for real-time electricity trading, bridging the DAM and BM by allowing continuous trading up to a few hours before delivery. This enables participants to adjust their positions in response to unforeseen events, such as plant outages or fluctuations in renewable energy output \cite{shinde2019literature}. In contrast to the DAM's hourly settlement intervals, the IDM allows trades in 30-minute blocks, notably in markets like the Irish Single Electricity Market.The IDM's real-time nature necessitates advanced forecasting models capable of accurately predicting short-term price movements. These challenges necessitate advanced forecasting models capable of managing the variability introduced by weather-dependent renewables \cite{piantadosi2024photovoltaic}, with explainable frameworks helping ensure adaptability and reliability in dynamic conditions \cite{samarajeewa2024artificial}. Traditional statistical methods \cite{uniejewski2019understanding, narajewski2020econometric} have shown success in processing large datasets and adapting to the market's dynamic conditions. Additionally, in markets with a high penetration of renewable energy, the IDM optimises trading strategies by enabling participants to refine their actions in near real-time, thereby mitigating risks associated with price volatility \cite{koch2019short, birkeland2024research}. Although the BM shares several structural similarities with the IDM—particularly in their real-time nature and focus on short-term trading—our work centres primarily on the BM due to its finer granularity and its role in balancing supply and demand in real-time.

Despite the importance of the BM in maintaining equilibrium between supply and demand, the literature on BM forecasting is relatively underdeveloped compared to that of the DAM and IDM. The BM plays a key role in balancing supply and demand with finer granularity than the IDM, particularly when responding to the fluctuations associated with renewable energy \cite{ortner2019future}. Accurate BM price forecasts are essential, especially where large-scale energy storage is not viable \cite{MAZZI2017259}. Studies have compared various forecasting models and their performance in BM settings, typically with limited horizons. Probabilistic forecasting methods for BM prices, such as those proposed by \cite{dumas2019probabilistic}, show promise but suggest further enhancements and integration into bidding strategies. Notably, existing research often lacks critical analysis and broader context. Recent studies, including \cite{narajewski2022probabilistic} and \cite{OCONNOR2024101436}, explore diverse models for BM price forecasting, with simpler models often outperforming the more complex ML and DL models. These studies highlight the need for continued research to address current limitations and explore emerging methodologies, ensuring sharp and reliable BM price predictions. Given the challenges of renewable energy integration and market volatility, European electricity market reforms increasingly prioritise resilient mechanisms \cite{zachmann2023design}.

\begin{table}[]
\begin{tabular}{p{4.3cm} p{1.0cm} p{1.5cm} p{1.0cm} p{1.4cm} p{1.5cm}}
\textbf{Reference} & \textbf{Point} & \textbf{Proba-bilistic} & \textbf{CP} & \textbf{Day-Ahead} &  \textbf{Real-Time}   \\

\cite{nowotarski2015computing}     &        &    \checkmark   &       &    \checkmark   &       \\
\cite{hagfors2016modeling}       &        &  \checkmark      &       &       &   \checkmark     \\
\cite{maciejowska2016probabilistic}&        &    \checkmark    &       &     \checkmark   &      \\
\cite{hagfors2016modeling} &        &    \checkmark   &       &       &   \checkmark    \\

\cite{lago2018forecasting}          &   \checkmark     &       &       &    \checkmark   &      \\
\cite{nowotarski2018recent}     &        &   \checkmark    &       &    \checkmark   &        \\
\cite{koch2019short}             &        &       &       &       &  \checkmark      \\
\cite{shinde2019literature}      &        &       &       &       &  \checkmark    \\
\cite{cheng2019hybrid}             &  \checkmark &       &       &  \checkmark &        \\
\cite{dumas2019probabilistic}      &        &    \checkmark   &       &       &    \checkmark     \\
\cite{uniejewski2019understanding} &    \checkmark    &       &       &       &   \checkmark    \\

\cite{narajewski2020econometric}     &        &   \checkmark    &       &       &   \checkmark     \\
\cite{lucas2020price}               &   \checkmark     &       &       &       &   \checkmark     \\
\cite{marcjasz2020probabilistic}     &        &   \checkmark    &       &   \checkmark    &      \\
\cite{lago2021forecasting}       &    \checkmark    &       &       &    \checkmark   &       \\
\cite{uniejewski2021regularized}     &        & \checkmark &       & \checkmark &            \\
\cite{kath2021conformal}             &        & \checkmark & \checkmark  &  \checkmark   & \checkmark   \\
\cite{narajewski2022probabilistic}   &        &   \checkmark    &       &       &    \checkmark   \\
\cite{khajeh2022applications}        &        &    \checkmark   &       &       &       \\
\cite{jensen2022ensemble}          &        &  \checkmark     &   \checkmark    &       &      \\
\cite{hu2022conformalized}      &        &    \checkmark    &    \checkmark    &       &      \\
\cite{zaffran2022adaptive}      &        &    \checkmark   &   \checkmark    &       &     \\

\cite{uniejewski2023smoothing}       &        & \checkmark &       &       &      \\
\cite{marcjasz2023distributional}  &        &   \checkmark    &       &     \checkmark  &     \\
\cite{subhankar2023probabilistically} &        &  \checkmark     &  \checkmark     &       &      \\

\cite{dewolf2023valid}          &        &   \checkmark    &   \checkmark    &       &       \\
\cite{trebbien2023probabilistic}        &        &   \checkmark    &       &   \checkmark   &      \\
\cite{cordier2023flexible}      &        &   \checkmark    &   \checkmark    &       &      \\

\end{tabular}
\end{table}

\begin{table}[]
\begin{tabular}{p{4.0cm} p{1.0cm} p{1.4cm} p{1.0cm} p{1.5cm} p{1.5cm}}
\textbf{Reference} & \textbf{Point} & \textbf{Probab-ilistic} & \textbf{CP} & \textbf{Day-Ahead} &  \textbf{Real-Time}   \\
\cite{OCONNOR2024101436}             &    \checkmark     &       &       &     \checkmark  &   \checkmark      \\
\cite{monjazeb2024wholesale}       &        &    \checkmark   &       &   \checkmark    &      \\

\cite{de2024electricity}        &        &   \checkmark    &   \checkmark    &    \checkmark   &      \\
\cite{lee2024transformer}       &        &    \checkmark    &     \checkmark   &       &       \\
\cite{xu2024conformal}          &        &     \checkmark  &   \checkmark    &       &       \\
\cite{angelopoulos2024conformal}&        &   \checkmark     \\
\textbf{Our approach} &       &   \checkmark    &   \checkmark    &   \checkmark    &    \checkmark      \\
\end{tabular}
\caption{Literature on PEPF. The \checkmark is the approach presented herein}
\label{TableLR}
\end{table}

\subsection{Probabilistic Methods}
Probabilistic forecasting methods have evolved substantially to meet the need for robust PIs in volatile electricity markets. Techniques such as Bayesian approaches, bootstrap methods, distribution-based forecasts, Monte Carlo simulations, and historical simulations each bring unique strengths but also face practical limitations. Bayesian methods effectively quantify uncertainty by combining prior knowledge with observed data, as seen in advancements like Bayesian neural networks (BNNs), stochastic volatility models, and hybrid Bayesian optimisation \cite{vahidinasab2010bayesian, kostrzewski2019probabilistic, cheng2019hybrid}. However, their high computational demands and reliance on strong priors limit scalability, particularly in non-stationary conditions. Bootstrap methods, such as residual-based bootstrapping, improve empirical coverage in BM settings but often produce overly wide intervals during volatile periods \cite{uniejewski2019importance, khosravi2013quantifying}. Distribution-based forecasts, including copula processes and generalised additive models, capture extreme price behaviours but are constrained by their reliance on correct distributional assumptions \cite{klein2023deep, serinaldi2011distributional, monteiro2018new}. Monte Carlo simulations, enhanced through Bayesian networks and sequential Markov Chain Monte Carlo techniques, improve real-time adaptability but remain computationally prohibitive for high-frequency trading \cite{yuanchen2023electricity}. Historical simulations offer simplicity and robustness but struggle to address structural shifts or unprecedented events, particularly in markets integrating renewable energy \cite{maciejowska2024multiple, janczura2024expectile}. Recent studies have explored the use of meta-heuristic optimization techniques for forecasting models in short-term load forecasting \cite{bhatnagar2025using}.

These limitations have driven interest in Conformal Prediction (CP) techniques, which offer a flexible, distribution-free framework for generating reliable PIs. CP avoids the restrictive assumptions of traditional methods while ensuring predefined confidence levels, even in non-stationary markets \cite{gammerman1998learning, vovk2005algorithmic}. Recent advancements, such as Ensemble Batch Prediction Intervals (EnbPI) and Sequential Predictive Conformal Inference (SPCI), extend CP’s utility to time-series applications \cite{xu2023sequential}. Despite these strengths, CP’s reliance on large calibration datasets and sensitivity to structural breaks can undermine performance in dynamic electricity markets \cite{foygel2022conformal}. Further research should focus on enhancing CP’s adaptability and robustness, particularly for handling extreme price events and evolving market complexities.

\subsubsection*{Quantile Regression \& Conformal Prediction}
QR and QRA are widely used in PEPF for their simplicity, efficiency, and ability to capture uncertainty across quantiles. Recent advancements have refined these techniques, such as Lasso QRA and smoothed QRA with kernel estimation, which improve predictive performance \cite{uniejewski2021regularized, uniejewski2023smoothing}. DL integrations have also shown promise, with methods like QR combined with LSTNet and SHAP-based feature selection enhancing sharpness through kernel density estimation \cite{liu2023day}, or hybrid QR-LSTM frameworks optimised via multi-objective algorithms improving accuracy under diverse market conditions \cite{xu2024novel}. While effective, QR and QRA face challenges in uncertainty quantification, particularly in achieving reliable coverage, motivating the exploration of CP as a complementary framework.

CP generates distribution-free PIs with predefined confidence levels, making it a robust alternative for uncertain and volatile markets \cite{gammerman1998learning, vovk2005algorithmic}. Extensions like Ensemble Batch Prediction Intervals (EnbPI), which combines ensemble learning with block bootstrap methods, have demonstrated success in addressing seasonality and trends in energy forecasting \cite{xu2021conformal, zaffran2022adaptive}. Similarly, Sequential Predictive Conformal Inference (SPCI) dynamically recalibrates PIs to improve reliability in non-stationary settings, with Transformer-based extensions further enhancing its adaptability \cite{lee2024transformer}. Hybrid approaches such as Ensemble Conformal QR and SCP with QR, offer a balance between computational efficiency and robust coverage, making them suitable for real-time applications \cite{jensen2022ensemble, cordier2023flexible}.  However, CP still faces practical limitations, including sensitivity to structural breaks and reliance on large calibration datasets \cite{foygel2022conformal}. Enhancing CP’s robustness, addressing extreme price events, and integrating CP with DL architectures to improve its accuracy and adaptability in dynamic electricity markets have proven beneficial \cite{angelopoulos2024conformal, xu2024conformal}.

\section{Methodology}\label{Meth}
This section outlines our study's methodology, emphasising the key models and forecasting approaches. It is structured as follows: Section \ref{modelss} introduces probabilistic forecasting models, followed by Section \ref{QRApproaches}, which describes the proposed uncertainty quantification approaches and our proposed ensemble, Q-Ens. 


\subsection{Electricity Price Prediction Models}\label{modelss}
The probabilistic regressor models benchmarked include:
\begin{itemize}
    \item \textit{LASSO Estimated AR} (LEAR): An autoregressive model with LASSO regularization for feature selection and handling high-dimensional data, providing probabilistic predictions through conditional distribution estimation.
    
    \item \textit{K-Nearest Neighbors} (KNN): A non-parametric model predicting outputs based on the "K" nearest neighbors, with probabilistic predictions derived from the neighbors' output distribution.
    
    \item \textit{Random Forest} (RF): An ensemble model combining multiple regression trees to reduce variance and maintain low bias, generating probabilistic predictions by aggregating tree outputs.
    
    \item \textit{Light Gradient Boosting Method} (LGBM): An ensemble model using boosting to reduce bias and variance, providing probabilistic forecasts by summing weighted outputs of weak learners (trees).
\end{itemize}

\subsection{Uncertainty Estimation}\label{QRApproaches}
In the domain of EPF, probabilistic forecasting methods offer a significant advancement over traditional point forecasts by providing insights into the uncertainty surrounding future prices.
In this section, we delve into various methods for PEPF in both the DAM and BM. We outline the approaches utilised, including QR, SCP, EnbPI, SPCI, QRA, and our proposed QR-CP Ensemble (Q-Ens). 

We frame the price prediction problem as a regression task. Given a dataset \(\mathcal{Z} = \{(\mathbf{x}_t, y_t)\}_{t=1}^N\), where \(\mathbf{x}_t\) represents the covariates and \(y_t\) the response variable, our objective at each time step \(t\) is to predict the next price \(y_t\) based on previous prices and covariates \(\mathbf{x}_t\).
The prediction is made using a model \(f\), such that:
\begin{equation}
\hat{y}_t = f(\mathbf{x}_t)
\end{equation}
where \(\hat{y}_t\) is the model's prediction. The models are trained on the historical dataset \(\mathcal{Z}\), which consists of \(N\) input-output pairs.
The conditional quantile function \( Q_{y_t}(\alpha \mid \mathbf{x}_t) \) estimates the value below which a certain percentage \(\alpha\) of the future price \(y_t\) is expected to fall, given the covariates \(\mathbf{x}_t\). We denote the lower quantile at level \(\alpha\) as \(\hat{q}_t^\alpha\) and the upper quantile at level \(1 - \alpha\) as \(\hat{q}_t^{1-\alpha}\). 
\begin{figure}[ht]
    \centering
    \begin{tikzpicture}
        \begin{axis}[
            width=11cm, 
            height=7cm,
            xlabel={Time},
            ylabel={\euro /MWh},
            title={},
            grid=both,
            xmin=0, xmax=12, 
            ymin=-1, ymax=34, 
            legend style={at={(0.01,0.99)}, anchor=north west}
        ]

        \addplot[blue, thick] coordinates {
            (0, 19) (1, 9) (2, 4) (3, 14) (4, 12) (5, 14) (6, 16) (7, 20) (8, 22) (9, 25) (10, 21) (11, 19) (12, 23)
        } ;
        \addlegendentry{Real Price}
        
        \addplot[blue, dashed, thick] coordinates {
            (8, 22) (9, 23) (10, 19) (11, 22) (12, 26)
        };
        \addlegendentry{Forecasted Price}
        
        \addplot[fill=gray, fill opacity=0.4] coordinates {
            (8, 20) (9, 20) (10, 15) (11, 17) (12, 20) 
            (12, 32) (11, 27) (10, 23) (9, 26) (8, 24) (8, 24)
        };
        \addlegendentry{Quantile Region}
        
        \node at (axis cs:8,20) [anchor=north] {$\hat{q}^{\alpha}_{t}$};
        \node at (axis cs:8,23) [anchor=south] {$\hat{q}^{1-\alpha}_{t}$};
        
        \node at (axis cs:9,19) [anchor=north] {$\hat{q}^{\alpha}_{t+1}$};
        \node at (axis cs:9,26) [anchor=south] {$\hat{q}^{1-\alpha}_{t+1}$};
        
        \node at (axis cs:10,15) [anchor=north] {$\hat{q}^{\alpha}_{t+2}$};
        \node at (axis cs:10,24) [anchor=south] {$\hat{q}^{1-\alpha}_{t+2}$};
        \node at (axis cs:11,16) [anchor=north] {$\hat{q}^{\alpha}_{t+3}$};
        \node at (axis cs:11,29) [anchor=south] {$\hat{q}^{1-\alpha}_{t+3}$};
        \node at (axis cs:12,18) [anchor=north] {$\hat{q}^{\alpha}_{t+4}$};
        \node at (axis cs:12,-100) [anchor=south] {$\hat{q}^{1-\alpha}_{t+4}$};    
        \end{axis}
    \end{tikzpicture}
    \caption{Quantile forecast of electricity prices}
    \label{QuantilePlot}
\end{figure}
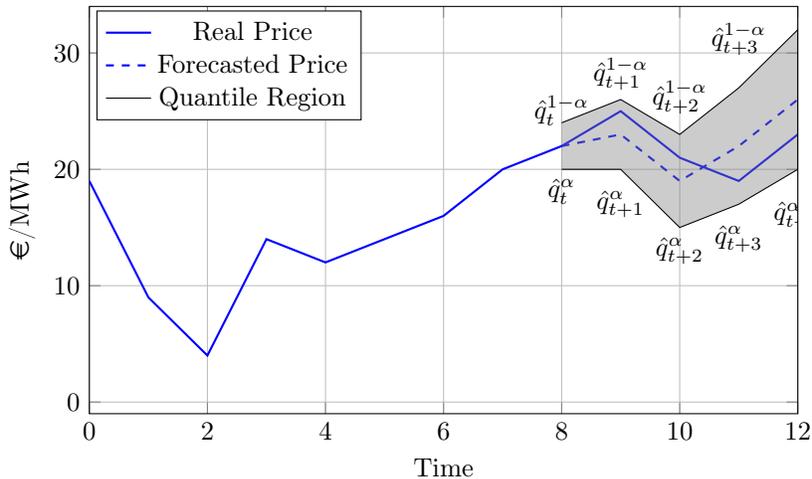
In the probabilistic prediction setting, as shown in Figure \ref{QuantilePlot}, the model \(f\) generates a PI defined by these quantiles:
\begin{equation}
\Gamma_{1-2\alpha}(\mathbf{x}_t) = [\hat{q}_t^\alpha, \hat{q}_t^{1-\alpha}]
\end{equation}
This interval indicates that the future price \(y_t\) is expected to fall within the range \([\hat{q}_t^\alpha, \hat{q}_t^{1-\alpha}]\) with a confidence level of \(1 - 2\alpha\). For example, if \(\alpha = 0.1\), the PI corresponds to an 80\% confidence level (\(1 - 2\alpha = 0.8\)), implying that there is an 80\% probability that \(y_t\) will lie within the interval.

\subsubsection{Quantile Regression}\label{QuantileRegression}
QR, introduced in \cite{koenker1978regression}, extends ordinary least squares regression to estimate conditional quantiles of the response variable rather than just the mean. QR provides a more comprehensive view by estimating the conditional median or other quantiles, revealing how covariates influence the entire distribution of the response variable.
Consider a linear model of the form:
\begin{equation}
y_t = \mathbf{x}_t^\top \mathbf{\beta} + \epsilon_t
\label{eq1}
\end{equation}
where $y_t$ is the dependent variable for the $t$-th observation, $\mathbf{x}_t$ is a $K$-dimensional vector of covariates, $\mathbf{\beta}$ is a $K$-dimensional vector of coefficients, and $\epsilon_t$ is the error term.

The goal of QR is to estimate the conditional quantile function, which represents the value below which a certain percentage \(\alpha\) of the response variable \(y_t\) falls, given the predictor variable \(\mathbf{x_{t}}\).
For a given $\alpha$ quantile level, the QR estimator $\hat{\beta}_\alpha$ is defined as the solution to the following optimisation problem:
\begin{equation}
\hat{\beta}_\alpha = \arg\min_{\beta} \sum_{t=1}^N \rho_\alpha (y_t - \mathbf{x}_t^\top \mathbf{\beta})
\label{eq2}
\end{equation}
where $\rho_\alpha(u)$ is the check function defined by:
\begin{equation}
\rho_\alpha(u) = u (\alpha - \mathbf{1}{\{u < 0\}})
\label{eq3}
\end{equation}

Here, $\mathbf{1}{\{u < 0\}}$ is an indicator function that equals 1 if $u < 0$ and 0 otherwise. The check function can be rewritten as:
\begin{equation}
\rho_\alpha(u) =
\begin{cases}
\alpha u & \text{if } u \ge 0 \\
(\alpha - 1)u & \text{if } u < 0
\end{cases}
\label{eq4}
\end{equation}
This function penalises underestimation and overestimation asymmetrically, depending on the value of $\alpha$. Specifically, it assigns a weight of $\alpha$ to positive residuals and $1 - \alpha$ to negative residuals, allowing the estimation of different quantiles.

The asymptotic properties of the QR estimator $\hat{\beta}_\alpha$ are well-established under suitable regularity conditions. The estimator $\hat{\beta}_\alpha$ is consistent, meaning it converges to the true parameter value as the sample size increases, and it is asymptotically normally distributed, approaching a normal distribution as the sample size grows. The asymptotic covariance matrix of $\hat{\beta}_\alpha$ is given by:
\begin{equation}
\text{Cov}(\hat{\beta}_\alpha) = \alpha (1 - \alpha) \left[ f_{y_t|\mathbf{x}_t}(\hat{q}_t^\alpha)^2 \cdot (X'X)^{-1} \right]
\label{eq5}
\end{equation}
where $\alpha (1 - \alpha)$ adjusts for the variability in estimating the quantile level $\alpha$, $f_{y_{t}|\mathbf{x}_t}(\hat{q}_t^\alpha)$ is the conditional density of $y_t$ given $\mathbf{x}_t$ at the $\alpha$ quantile level, reflecting the density around this quantile, and $(X'X)^{-1}$ is the inverse of the product of the design matrix with its transpose. The design matrix $X$ contains the values of all predictor variables for each observation $X =  \{ (\mathbf{x}_t)\}_{t=1}^N$, and its inverse accounts for the scaling and correlations among these predictors.

\subsubsection*{Conformal Prediction}
CP is a flexible, distribution-free framework for quantifying the uncertainty of predictions made by machine learning models. CP generates prediction sets with a predefined confidence level, ensuring that the true value lies within these sets with a specified probability.
At the core of CP is the concept of nonconformity, which measures how unusual a new data point \((\mathbf{x}_{\text{new}}, y_{\text{new}})\) is compared to a set of previously observed examples. A nonconformity measure \(A\) is defined to assign a nonconformity score to each data point in the training set \(\mathcal{Z}\).
Traditional CP methods use p-values calculated from nonconformity scores to construct prediction sets. For a new example \(\mathbf{x}_{\text{new}}\) with a candidate label \(y_{\text{new}}\), the p-value is given by:
$p(y_{\text{new}}) = $
\begin{equation}
\frac{|\{(\mathbf{x}_t, y_t) \in \mathcal{Z} : A(\mathbf{x}_t, y_t) \geq A(\mathbf{x}_{\text{new}}, y_{\text{new}})\}| + 1}{N + 1}
\end{equation}
where \(N\) is the size of \(\mathcal{Z}\). The prediction set for \(\mathbf{x}_{\text{new}}\) includes all candidate labels for which the p-value exceeds a certain significance level.

\subsubsection{Split Conformal Prediction}\label{SplitConformalPrediction}
SCP, also known as inductive CP, is a simplified variation of standard conformal prediction that enhances computational efficiency by dividing the data into two separate sets: a training set and a calibration set. The training set is used to fit a predictive model, while the calibration set is used to generate nonconformity scores, which measure the discrepancy between the true and predicted values.
The data is split into two subsets: a training set $\mathcal{Z}_{\text{train}}$ and a calibration set $\mathcal{Z}_{\text{cal}}$. A predictive model $f$ is trained on $\mathcal{Z}_{\text{train}}$, and the model's performance is assessed on the calibration set by computing nonconformity scores for each pair $(\mathbf{x}_t, y_t) \in \mathcal{Z}_{\text{cal}}$.

The nonconformity score for an observation $t$ is defined as:
\begin{equation}
A(\mathbf{x}_t, y_t) = |y_t - f(\mathbf{x}_t)|    
\label{eq:residual}
\end{equation}
This score measures the absolute difference between the true value $y_t$ and the predicted value $f(\mathbf{x}_t)$, with larger scores indicating higher nonconformity between the model's prediction and the observed outcome.
Once nonconformity scores are computed for all calibration examples, a quantile of the nonconformity scores is selected to construct PIs for new data points. Let $\lambda_t$ be the quantile for time period $t$, corresponding to the desired confidence level $1 - 2\alpha$. This ensures that at least $1 - 2\alpha$ of the calibration examples have nonconformity scores smaller than or equal to $\lambda_t$, guaranteeing that the PIs will cover the true value with confidence level $1 - 2\alpha$.

For a new example $\mathbf{x}_{t}$, the PI is constructed as:
\begin{equation}
\Gamma_{1-2\alpha}(\mathbf{x}_{t}) = \left[f(\mathbf{x}_{t}) - \lambda_t, f(\mathbf{x}_{t}) + \lambda_t\right]
\end{equation}
This interval represents the range of possible values for the response variable $y_{t}$ that are consistent with the nonconformity scores computed from the calibration set. 
SCP balances computational efficiency and prediction robustness by calculating nonconformity scores on a smaller calibration set, avoiding the need to recompute predictions for each test example. This reduces the computational load but assumes data exchangeability, which limits its applicability in settings like time series data.

\subsubsection{Ensemble Batch Predictive Interval (EnbPI)}\label{EnbPI}
EnbPI, introduced in \cite{xu2021conformal}, is a probabilistic method within the CP framework, specifically designed to generate PIs with strong coverage guarantees. EnbPI leverages ensemble learning techniques and bootstrap sampling to provide reliable PIs, particularly for forecasting applications. Unlike traditional CP methods that require a separate calibration set, EnbPI adopts a leave-one-out approach, which enables the method to maintain valid coverage without sacrificing data for calibration.
Given a dataset \(\mathcal{Z}\), EnbPI constructs PIs by first generating \(B\) bootstrap samples from the original dataset. A predictive model \(f^{(b)}\) is trained on each bootstrap sample \(\mathcal{Z}^{(b)}\), allowing EnbPI to account for model variability and provide more reliable PIs.
For each trained model \(f^{(b)}\), residuals are computed for every observation \((\mathbf{x}_t, y_t)\) by calculating the absolute difference between the observed value \(y_t\) and the predicted value \(f^{(b)}(\mathbf{x}_t)\). The residuals are given by:
\begin{equation}
A^{(b)}(\mathbf{x}_t, y_t) = |y_t - f^{(b)}(\mathbf{x}_t)|
\end{equation}

These residuals reflect the model’s prediction errors and are used to determine the width of the PIs. The quantile \(\lambda_{t}^{(b)}\) for each bootstrap model is defined as:
\begin{equation}
\lambda_{t}^{(b)} = \text{Quantile}_{1-2\alpha}(A_1^{(b)}, A_2^{(b)}, \ldots, A_N^{(b)})
\end{equation}

The PI for a new input \(\mathbf{x}_{t}\) is constructed as $\Gamma_{1-2\alpha}(\mathbf{x}_{t})$, for a confidence level of $1-2\alpha$, by averaging the predictions from all bootstrap models and incorporating the margin \(\lambda_t\). The resulting PI is given by:
\begin{equation}
\Gamma_{1-2\alpha}(\mathbf{x}_{t}) = \left[ \frac{1}{B} \sum_{b=1}^B f^{(b)}(\mathbf{x}_{t}) - \lambda_t, \frac{1}{B} \sum_{b=1}^B f^{(b)}(\mathbf{x}_{t}) + \lambda_t \right]
\end{equation}
where \(\lambda_t\) is the average of the quantiles \(\lambda_{t}^{(b)}\) across all bootstrap models:
\begin{equation}
\lambda_t = \frac{1}{B} \sum_{b=1}^B \lambda_{t}^{(b)}
\end{equation}
By averaging across multiple models and adjusting the interval width using the quantile of residuals, EnbPI mitigates the effects of outliers and reduces the variance in PIs. 
Its flexibility allows use across various models and data types, making it ideal for complex EPF. The use of ensemble learning also improves the resilience of the PIs.

\subsubsection{Sequential Predictive Conformal Inference (SPCI)}\label{SPCI}
SPCI, introduced in \cite{xu2023sequential}, extends the EnbPI method to handle non-exchangeable time series data. Recognizing the limitations of traditional CP methods in the context of temporal dependencies, SPCI incorporates quantile regression forests (QRF) to estimate conditional quantiles of residuals. This allows SPCI to better adapt to the temporal dependencies and heteroskedasticity often encountered in time series data. 
SPCI works by training a predictive model \(f\) on the dataset \(\mathcal{Z} \). The model’s residuals are computed using Equation (\ref{eq:residual}) on a rolling calibration set and used to estimate future residuals through QR. By applying QRF to the residuals, SPCI estimates the conditional quantile $\hat{q}_t^\alpha$, where \(1 - 2\alpha\) represents the desired confidence level. This approach helps SPCI account for the serial dependence in the residuals, which is critical for accurate PIs in time series forecasting.
To generate PIs for a new input \(\mathbf{x}_{t}\), SPCI constructs the interval:
\begin{equation}
\Gamma_{1-2\alpha}(\mathbf{x}_{t}) = \left[f(\mathbf{x}_{t}) - \lambda_t, f(\mathbf{x}_{t}) + \lambda_t\right]
\end{equation}
where \(\lambda_t\) is derived from the quantiles of past residuals, adjusted for the level of uncertainty in the data.
One of the main advantages of SPCI is its ability to dynamically re-estimate residual quantiles as new data becomes available. This adaptive mechanism allows the method to maintain valid coverage over time, ensuring that the true outcome falls within the PI with a specified confidence level ($1-2\alpha$). Furthermore, SPCI is applicable to a wide range of predictive models and data types, making it highly versatile for complex time series forecasting tasks. Compared to SCP and EnbPI, SPCI often yields narrower intervals without sacrificing coverage, offering a powerful tool for uncertainty quantification in time series.

\subsubsection{Quantile Regression Averaging}
QRA was introduced in \cite{nowotarski2015computing}. QRA applies QR to a set of point forecasts from different models, allowing direct estimation of electricity spot price distribution without separating the probabilistic forecast into a point forecast and an error distribution. This method directly leverages electricity price distribution properties for improved forecasting accuracy.
It considers $M$ different point prediction models $f^m$ and feeds their output to a probabilistic model:
\begin{equation}
\Gamma_{1-2\alpha}(\mathbf{x}_t, [f^1(\mathbf{x}_t), f^2(\mathbf{x}_t), \dots, f^M(\mathbf{x}_t)]) = [\hat{q}_t^\alpha, \hat{q}_t^{1-\alpha}]
\end{equation}

In our implementation, two sets of point forecasts are generated:
\begin{itemize}
    \item \underline{QRA-R:} Includes forecasts from models trained on different data contexts (full datasets, historical, and future datasets) to capture diverse trends and improve reliability.
    \item \underline{QRA-CP:} Consists of forecasts from various Conformal Prediction (CP) models like SCP, EnbPI, and SPCI, each contributing unique strengths to quantile predictions.
\end{itemize}

\subsubsection{QR-CP Ensemble: A Robust Approach}\label{QRAQR}
In predictive modeling, uncertainty is categorised into epistemic and aleatoric types. Epistemic uncertainty arises from limited knowledge of the underlying model and can be reduced with more data or model improvements, while aleatoric uncertainty results from inherent variability in the data and cannot be mitigated by additional data. Our Q-Ens combines QR, EnbPI, and SPCI aims to address both uncertainties, improving the reliability and sharpness for PEPF.
QR captures a wide range of possible outcomes by estimating multiple conditional quantiles, thus addressing both epistemic and aleatoric uncertainties. EnbPI reduces epistemic uncertainty by averaging predictions from bootstrap samples, thereby lowering variance and reflecting aleatoric uncertainty in the constructed PIs. SPCI adapts dynamically to new data, updating PIs in real-time to account for evolving data variability, thus capturing both types of uncertainty as they change over time.

\begin{figure}
\centerline{\includegraphics[width=1\linewidth]{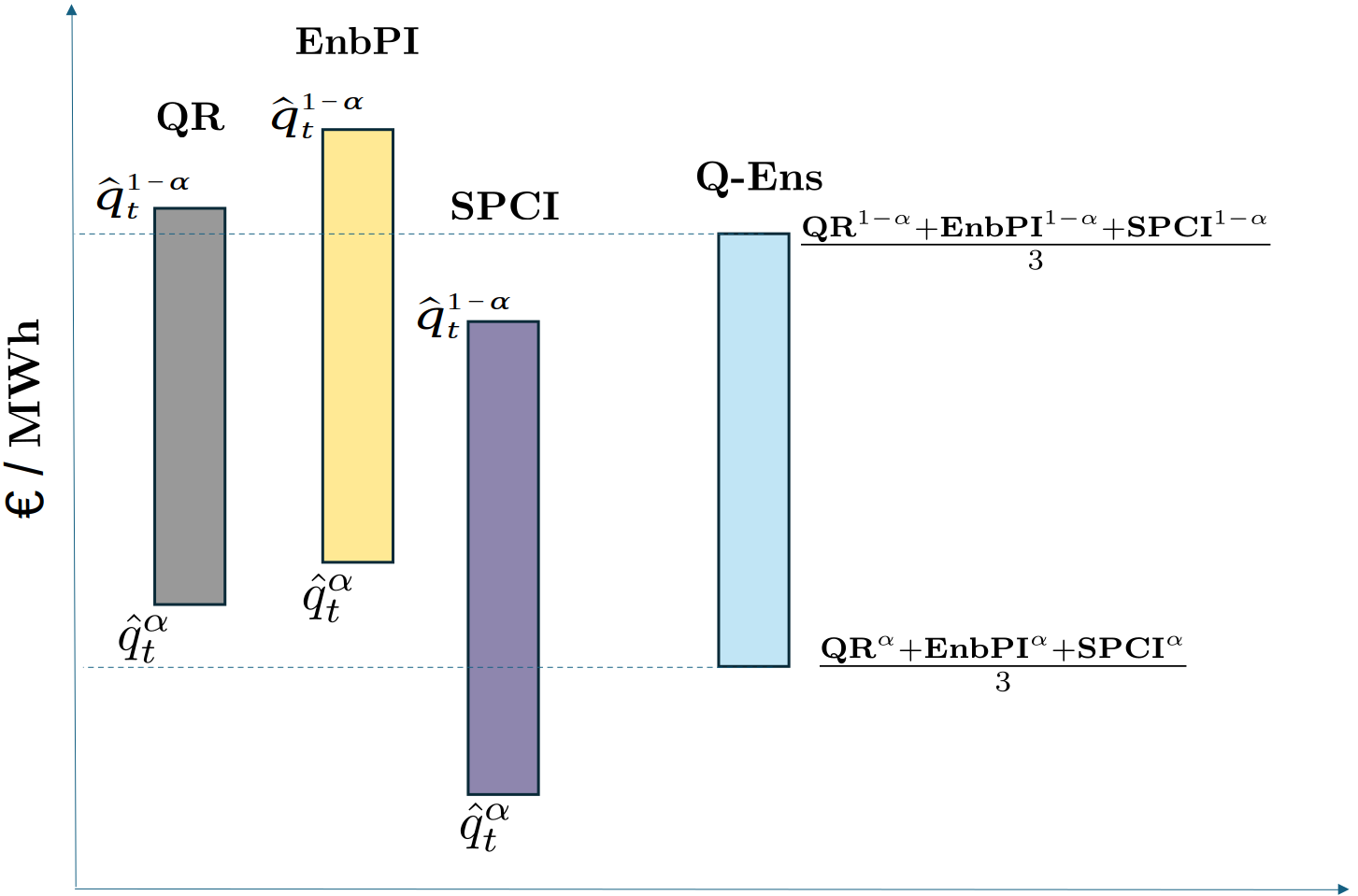}}
\caption{Construction of the Ensemble, Q-Ens.}
\label{QEnsConstruction}
\end{figure}

The ensemble combines the strengths of these methods, working from a similar approach as EnbPI, averaging the models outputs to both reduce epistemic uncertainty further, and improve overall forecast performance. The ensemble prediction is obtained by averaging the quantile predictions from QR, EnbPI, and SPCI models:
$\Gamma_{1-2\alpha} =$
\begin{equation}
[\frac{1}{3} (\text{QR}^{\alpha} + \text{EnbPI}^{\alpha} + \text{SPCI}^{\alpha}), \frac{1}{3} (\text{QR}^{1-\alpha} + \text{EnbPI}^{1-\alpha} + \text{SPCI}^{1-\alpha})]
\end{equation}
with an outline of the PIs construction shown in Figure \ref{QEnsConstruction}.

\section{Evaluation Metrics}\label{EvalMetrics}
Our evaluation of probabilistic predictions hinges on two crucial dimensions: validity and efficiency. Efficiency, gauged through metrics like sharpness and interval width, enhances precision. Simultaneously, validity, including calibration and coverage, requires the generation of precise PIs to affirm the reliability of our forecasts. 

\subsection{Efficiency}
Efficiency in probabilistic forecasting is crucial for precision. It is evaluated using two key metrics: the Pinball Score and Interval Width.

\subsubsection{Pinball Score}\label{sharpness}
In both the DAM and BM, sharpness plays an important role in accurate anticipation, hedging, and real-time adjustments. The Pinball Score \cite{zhao2019adaptive}, derived from the Pinball Loss function, measures the sharpness of the forecast based on the quantile prediction $\hat{q}^{\alpha}_{t}$ and the observed price $y_{t}$ for time period $t$:
\begin{align}
PS(\hat{q}^{\alpha}_{t}, y_{t}, \alpha) =
\begin{cases}
(1 - \alpha)(\hat{q}^{\alpha}_{t} - y_{t}) & \text{for } y_{t} \le \hat{q}^{\alpha}_{t}  \\ 
\alpha (y_{t} - \hat{q}^{\alpha}_{t}) & \text{for } y_{t} \geq \hat{q}^{\alpha}_{t}
\end{cases}
\end{align}
Here, $\alpha$ denotes the quantile level, such as 0.1 for the 10th percentile; hence, an 80\% PI ($[\hat{q}_t^\alpha, \hat{q}_t^{1-\alpha}]$) corresponds to quantile levels 0.1 and 0.9.


\subsubsection{Interval Width}
The Interval Width is another important metric for assessing the efficiency of probabilistic forecasts. While sharpness ensures reliability, efficiency is pivotal in refining precision, and it is intricately linked to the width of PIs. The Interval Width ($IW_{t}$) for a given PI can be defined as follows:
\begin{align}\label{eqIW}
IW_{t} = \hat{q}^{1-\alpha}_{t}- \hat{q}^{\alpha}_{t}
\end{align}
where $\hat{q}^{1-\alpha}_{t}$ and $\hat{q}^{\alpha}_{t}$ are the upper and lower quantiles of the PI, respectively. Narrower intervals indicate higher efficiency, reflecting greater precision in the forecast.

\subsection{Validity}
Validity in probabilistic forecasting ensures that the predicted intervals accurately reflect the true distribution of the observed data. Two primary metrics used to evaluate validity are coverage and the Winkler Score.

\subsubsection{Coverage}
Both reliability and accuracy of probabilistic forecasts are crucial, with a particular focus on evaluating their validity. We examine the models' coverage to highlight their ability to capture the true distribution, focusing on precision in capturing the price $y_t$ within a predefined PIs. The empirical coverage, indicating the agreement of predictions with the specified intervals, is denoted by the binary indicator $I_{t}^\alpha$:
\begin{align}
I_{t}^\alpha= \begin{cases}
1 & \text{if } y_{t} \in [\hat{q}^{\alpha}_{t}, \hat{q}^{1-\alpha}_{t}] \\
0 & \text{if } y_{t} \notin [\hat{q}^{\alpha}_{t}, \hat{q}^{1-\alpha}_{t}]
\end{cases}
\end{align}
The Coverage Probability ($\Lambda$) is then defined as:
\begin{align}
\Lambda = \frac{1}{N} \sum_{t}^N I_{t}^\alpha
\end{align}
where $N$ is the total number of forecasts. Ideally, the coverage probability should be close to the nominal level (e.g., 80\% if $\alpha$ =0.1), indicating that the PIs are valid.

\subsubsection{Winkler Score}
The Winkler Score ($W_{t}$) combines reliability and sharpness into a single metric, offering a comprehensive assessment of probabilistic forecasts.
This score evaluates the interval width ($IW_{t}$), see equation \ref{eqIW}, of PIs based on the observed values ($y_{t}$), incorporating a penalty factor ($\tau$) for deviations from the interval bounds:
\[
W_{t} = 
\begin{cases} 
IW_{t} & \text{if } y_{t} \in [\hat{q}^{\alpha}_{t}, \hat{q}^{1-\alpha}_{t}] \\
IW_{t} + \frac{2}{\tau} (\hat{q}^{\alpha}_{t} - y_{t}) & \text{if } y_{t} < \hat{q}^{\alpha}_{t} \\
IW_{t} + \frac{2}{\tau} (y_{t} - \hat{q}^{1-\alpha}_{t}) & \text{if } y_{t} > \hat{q}^{1-\alpha}_{t} 
\end{cases}
\]
The Winkler Score penalises larger deviations when observed prices fall outside the PIs, offering insight into the model's balance of accuracy and interval width.

\section{Case Studies}\label{resultssec}
We present two case studies to demonstrate the effectiveness of our QR-CP Ensemble (Q-Ens) approach for PEPF in the DAM and BM. Utilizing datasets from \cite{OCONNOR2024101436} spanning 2019 to 2022, we focus on the Irish electricity market, which is characterized by high curtailment rates due to surplus renewable generation during low-demand periods \cite{yasuda2022curtailment}. 
This section evaluates probabilistic approaches and forecasting models in both the DAM \ref{DAM_Eval} and the BM \ref{BM_Eval}. We examine the effectiveness of each model, looking at QR and CP approaches, with a focus on the performance of our proposed QR-CP Ensemble. The analysis can be divided into two main aspects: efficiency and validity. Efficiency evaluates the sharpness of the forecasts by examining metrics such as the Aggregate Pinball Score (APS) and interval width. Validity assesses the reliability of the forecasts by measuring each approach's ability to achieve coverage. These two aspects are integrated using the Winkler Score, which ties together efficiency and validity. This culminates in a financial performance evaluation of these probabilistic methods in Section \ref{financialperf}. To ensure reproducibility, the Python code and dataset used for this section are available on \href{https://anonymous.4open.science/r/PEPF_Conformal-C0AF/}{GitHub}\footnote{\url{https://anonymous.4open.science/r/PEPF_Conformal-C0AF/}}.

\subsection{Day-Ahead Market}\label{DAM_Eval}
The DAM dataset includes historical data and the system operator (TSO) predictions, revealing past market behaviors and aiding in forecasting future conditions.
In this case study, we predict DAM prices for the next 24 settlement periods using historical DAM prices and TSO wind and demand forecasts from the past 168 hours.

\subsubsection{Efficiency in the Day-Ahead Market: Pinball Score \&  Interval Width}\label{sharpnessDAM}
Analyzing the APS for DAM models in Table \ref{APS_DAM} highlights the strong performance for the Q-Ens approach, producing the two lowest APS in the DAM for the RF and LGBM  models. 
\begin{table}[ht!]
    \centering
    \setlength{\tabcolsep}{4pt} 
    \begin{tabular}{l r r r r r r r r}
        \toprule
         Model & \multicolumn{1}{c}{QR} & \multicolumn{1}{c}{SCP} & \multicolumn{1}{c}{EnbPI} & \multicolumn{1}{c}{SPCI} & \multicolumn{1}{c}{QRA-R} &  \multicolumn{1}{c}{QRA-CP} &  \multicolumn{1}{c}{Q-Ens} \\
        \midrule
        KNN  & 6.65 & \cellcolor{red!20}7.79 & \textbf{5.71} & 6.09 & 6.20  & 6.01 & 5.96 \\
        LEAR & 8.35 & 4.81 & 4.18 & \textbf{4.08} & \cellcolor{red!20}8.36 & 8.35 & 4.54 \\
        LGBM & 3.62 & 3.67 & \cellcolor{red!20}3.76 & 3.71 & 3.67 & 3.66 & \textbf{3.48} \\
        RF   & 3.63 & \cellcolor{red!20}3.85 & 3.81 & 3.84 & 3.60  & 3.61 & \textbf{3.59} \\
        Avg. & \cellcolor{red!20}5.56 & 5.03 & \textbf{4.37} & 4.43 & 5.43 & 5.36 & 4.39 \\
        \bottomrule
    \end{tabular}
    \caption{APS for the DAM. Best approach for each regressor marked in bold, worst in red.}
    \label{APS_DAM}
\end{table}
This includes strong performance for CP approaches EnbPI and SPCI, in particular improving LEAR and KNN models, taking advantage of CPs model agnostic feature. Time series adapted CP approaches EnbPI and SPCI also demonstrate notable improvements over SCP.
QR and QRA models struggle, but this is largely as a result of their poor performance with LEAR, substantially dragging down the average.

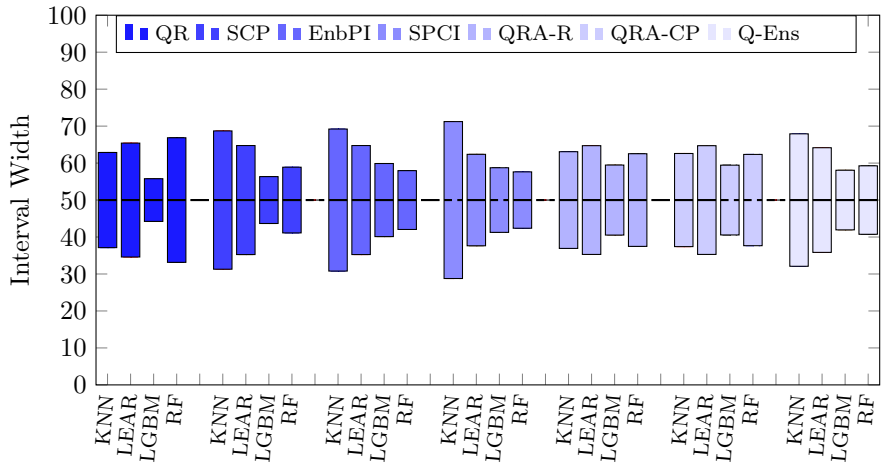
\begin{figure}[ht!]
        \centering
\begin{tikzpicture}
\begin{axis}[
    samples=101,
    width=12.0cm,
    height=6.5cm,
    boxplot/draw direction=y,
    ylabel={Interval Width},
    ytick={0,10, 20, 30, 40, 50, 60, 70, 80, 90, 100}, 
    ymin=0, 
    ymax=100, 
    xtick={1,2,3,4,5,6,7,8,9,10,11,12,13,14,15,16, 17, 18, 19, 20, 21, 22, 23, 24, 25, 26, 27, 28, 29, 30, 31, 32, 33, 34},
    xticklabels={KNN, LEAR, LGBM, RF, , KNN, LEAR, LGBM, RF, , KNN, LEAR, LGBM, RF, , KNN, LEAR, LGBM, RF, ,KNN, LEAR, LGBM, RF, , KNN, LEAR, LGBM, RF, , KNN, LEAR, LGBM, RF},
    xticklabel style={rotate=85, anchor=east, align=left, font=\footnotesize}, 
    boxplot/every box/.style={
        draw=black, 
        solid  
    },
    boxplot/every median/.style={
        black, 
        thick  
    },
    xmin=0.5, 
    xmax=34.5  
]

\addplot+ [
    boxplot prepared={
        draw position=1,
        lower whisker=37.13,
        lower quartile=37.13,
        median=50,
        upper quartile=62.87,
        upper whisker=62.87,
    },
    fill=blue!90
] coordinates {};

\addplot+ [
    boxplot prepared={
        draw position=2,
        lower whisker=34.595,
        lower quartile=34.595,
        median=50,
        upper quartile=65.405,
        upper whisker=65.405,
    },
    fill=blue!90,
    solid  
] coordinates {};

\addplot+ [
    boxplot prepared={
        draw position=3,
        lower whisker=44.225,
        lower quartile=44.225,
        median=50,
        upper quartile=55.775,
        upper whisker=55.775,
    },
    fill=blue!90,
    solid  
] coordinates {};

\addplot+ [
    boxplot prepared={
        draw position=4,
        lower whisker=33.15,
        lower quartile=33.15,
        median=50,
        upper quartile=66.85,
        upper whisker=66.85,
    },
    fill=blue!90,
    solid  
] coordinates {};

\addplot+ [
    boxplot prepared={
        draw position=5,
        lower whisker=50,
        lower quartile=50,
        median=50,
        upper quartile=50,
        upper whisker=50,
    },
    fill=blue!20
] coordinates {};

\addplot+ [
    boxplot prepared={
        draw position=6,
        lower whisker=31.295,
        lower quartile=31.295,
        median=50,
        upper quartile=68.705,
        upper whisker=68.705,
    },
    fill=blue!75,
    solid  
] coordinates {};

\addplot+ [
    boxplot prepared={
        draw position=7,
        lower whisker=35.255,
        lower quartile=35.255,
        median=50,
        upper quartile=64.745,
        upper whisker=64.745,
    },
    fill=blue!75,
    solid  
] coordinates {};

\addplot+ [
    boxplot prepared={
        draw position=8,
        lower whisker=43.66,
        lower quartile=43.66,
        median=50,
        upper quartile=56.34,
        upper whisker=56.34,
    },
    fill=blue!75,
    solid  
] coordinates {};

\addplot+ [
    boxplot prepared={
        draw position=9,
        lower whisker=41.11,
        lower quartile=41.11,
        median=50,
        upper quartile=58.89,
        upper whisker=58.89,
    },
    fill=blue!75,
    solid  
] coordinates {};

\addplot+ [
    boxplot prepared={
        draw position=10,
        lower whisker=50,
        lower quartile=50,
        median=50,
        upper quartile=50,
        upper whisker=50,
    },
    fill=blue!20
] coordinates {};

\addplot+ [
    boxplot prepared={
        draw position=11,
        lower whisker=30.79,
        lower quartile=30.79,
        median=50,
        upper quartile=69.21,
        upper whisker=69.21,
    },
    fill=blue!60,
    solid  
] coordinates {};

\addplot+ [
    boxplot prepared={
        draw position=12,
        lower whisker=35.255,
        lower quartile=35.255,
        median=50,
        upper quartile=64.745,
        upper whisker=64.745,
    },
    fill=blue!60,
    solid  
] coordinates {};

\addplot+ [
    boxplot prepared={
        draw position=13,
        lower whisker=40.115,
        lower quartile=40.115,
        median=50,
        upper quartile=59.885,
        upper whisker=59.885,
    },
    fill=blue!60
] coordinates {};

\addplot+ [
    boxplot prepared={
        draw position=14,
        lower whisker=42.045,
        lower quartile=42.045,
        median=50,
        upper quartile=57.955,
        upper whisker=57.955,
    },
    fill=blue!60
] coordinates {};

\addplot+ [
    boxplot prepared={
        draw position=15,
        lower whisker=50,
        lower quartile=50,
        median=50,
        upper quartile=50,
        upper whisker=50,
    },
    fill=blue!20
] coordinates {};

\addplot+ [
    boxplot prepared={
        draw position=16,
        lower whisker=28.775,
        lower quartile=28.775,
        median=50,
        upper quartile=71.225,
        upper whisker=71.225,
    },
    fill=blue!45
] coordinates {};

\addplot+ [
    boxplot prepared={
        draw position=17,
        lower whisker=37.63,
        lower quartile=37.63,
        median=50,
        upper quartile=62.37,
        upper whisker=62.37,
    },
    fill=blue!45,
    solid  
] coordinates {};

\addplot+ [
    boxplot prepared={
        draw position=18,
        lower whisker=41.26,
        lower quartile=41.26,
        median=50,
        upper quartile=58.74,
        upper whisker=58.74,
    },
    fill=blue!45
] coordinates {};

\addplot+ [
    boxplot prepared={
        draw position=19,
        lower whisker=42.375,
        lower quartile=42.375,
        median=50,
        upper quartile=57.625,
        upper whisker=57.625,
    },
    fill=blue!45,
    solid  
] coordinates {};

\addplot+ [
    boxplot prepared={
        draw position=20,
        lower whisker=50,
        lower quartile=50,
        median=50,
        upper quartile=50,
        upper whisker=50,
    },
    fill=blue!20
] coordinates {};

\addplot+ [
    boxplot prepared={
        draw position=21,
        lower whisker=36.92,
        lower quartile=36.92,
        median=50,
        upper quartile=63.08,
        upper whisker=63.08,
    },
    fill=blue!30,
    solid  
] coordinates {};

\addplot+ [
    boxplot prepared={
        draw position=22,
        lower whisker=35.29,
        lower quartile=35.29,
        median=50,
        upper quartile=64.71,
        upper whisker=64.71,
    },
    fill=blue!30,
    solid  
] coordinates {};

\addplot+ [
    boxplot prepared={
        draw position=23,
        lower whisker=40.53,
        lower quartile=40.53,
        median=50,
        upper quartile=59.47,
        upper whisker=59.47,
    },
    fill=blue!30,
    solid  
] coordinates {};

\addplot+ [
    boxplot prepared={
        draw position=24,
        lower whisker=37.47,
        lower quartile=37.47,
        median=50,
        upper quartile=62.53,
        upper whisker=62.53,
    },
    fill=blue!30,
    solid  
] coordinates {};

\addplot+ [
    boxplot prepared={
        draw position=25,
        lower whisker=50,
        lower quartile=50,
        median=50,
        upper quartile=50,
        upper whisker=50,
    },
    fill=blue!20
] coordinates {};

\addplot+ [
    boxplot prepared={
        draw position=26,
        lower whisker=37.405,
        lower quartile=37.405,
        median=50,
        upper quartile=62.595,
        upper whisker=62.595,
    },
    fill=blue!20,
    solid  
] coordinates {};

\addplot+ [
    boxplot prepared={
        draw position=27,
        lower whisker=35.295,
        lower quartile=35.295,
        median=50,
        upper quartile=64.705,
        upper whisker=64.705,
    },
    fill=blue!20,
    solid  
] coordinates {};

\addplot+ [
    boxplot prepared={
        draw position=28,
        lower whisker=40.56,
        lower quartile=40.56,
        median=50,
        upper quartile=59.44,
        upper whisker=59.44,
    },
    fill=blue!20
] coordinates {};

\addplot+ [
    boxplot prepared={
        draw position=29,
        lower whisker=37.655,
        lower quartile=37.655,
        median=50,
        upper quartile=62.345,
        upper whisker=62.345,
    },
    fill=blue!20
] coordinates {};

\addplot+ [
    boxplot prepared={
        draw position=30,
        lower whisker=50,
        lower quartile=50,
        median=50,
        upper quartile=50,
        upper whisker=50,
    },
    fill=blue!10
] coordinates {};

\addplot+ [
    boxplot prepared={
        draw position=31,
        lower whisker=32.085,
        lower quartile=32.085,
        median=50,
        upper quartile=67.915,
        upper whisker=67.915,
    },
    fill=blue!10
] coordinates {};

\addplot+ [
    boxplot prepared={
        draw position=32,
        lower whisker=35.845,
        lower quartile=35.845,
        median=50,
        upper quartile=64.155,
        upper whisker=64.155,
    },
    fill=blue!10,
    solid  
] coordinates {};

\addplot+ [
    boxplot prepared={
        draw position=33,
        lower whisker=41.935,
        lower quartile=41.935,
        median=50,
        upper quartile=58.065,
        upper whisker=58.065,
    },
    fill=blue!10
] coordinates {};

\addplot+ [
    boxplot prepared={
        draw position=34,
        lower whisker=40.74,
        lower quartile=40.74,
        median=50,
        upper quartile=59.26,
        upper whisker=59.26,
    },
    fill=blue!10,
    solid  
] coordinates {};

\node[anchor=north east, font=\footnotesize, fill=white, draw=black, text width=9.6cm, align=left] 
    at (axis cs:33.50, 100) {
        \parbox{14.0cm}{%
            \textcolor{blue!90}{\rule{0.1cm}{0.2cm}} \textcolor{blue!90}{\rule{0.1cm}{0.15cm}} QR 
            \textcolor{blue!75}{\rule{0.1cm}{0.2cm}} \textcolor{blue!75}{\rule{0.1cm}{0.15cm}} SCP
            \textcolor{blue!60}{\rule{0.1cm}{0.2cm}} \textcolor{blue!60}{\rule{0.1cm}{0.15cm}} EnbPI
            \textcolor{blue!45}{\rule{0.1cm}{0.2cm}} \textcolor{blue!45}{\rule{0.1cm}{0.15cm}} SPCI 
            \textcolor{blue!30}{\rule{0.1cm}{0.2cm}} \textcolor{blue!30}{\rule{0.1cm}{0.15cm}} QRA-R 
            \textcolor{blue!20}{\rule{0.1cm}{0.2cm}} \textcolor{blue!20}{\rule{0.1cm}{0.15cm}} QRA-CP 
            \textcolor{blue!10}{\rule{0.1cm}{0.2cm}} \textcolor{blue!10}{\rule{0.1cm}{0.15cm}} Q-Ens 
        }
    };

\end{axis}
\end{tikzpicture}
\caption{Interval Width for each model in the DAM}
\label{IWDAM}
\end{figure}

Looking further into efficiency, we evaluate the interval width of the models for the DAM in Figure \ref{IWDAM}, centered around a median of 50 for clarity. The analysis of IW across different models reveals key insights into model efficiency and precision. 
A pronounced contrasting outcome in Figure \ref{IWDAM} between QR and CP approaches is how CP approaches, SCP, EnbPI, and SPCI, result in a wider interval width for less accurate models KNN and LEAR, and a narrowing interval width for the more accurate RF and LGBM. SPCI, developed as a successor to EnbPI to produce narrower intervals, fails to achieve this for KNN in the DAM but succeeds slightly for LEAR, LGBM, and RF. This trend is absent entirely for QR and QRA approaches with QR and QRA producing intervals independently of models accuracy, with QRA-CP producing similarly sized Interval widths compared to QRA-QR, despite the contrasting contents of each approaches dataset.
Q-Ens closely mirrors EnbPI and SPCI for LEAR and KNN models. However, for LGBM, Q-Ens produces narrower interval widths due to the model's inherently narrow intervals when using QR. Conversely, Q-Ens with RF produces wider intervals than EnbPI or SPCI due to the wider interval width of QR.

\subsubsection{Validity in the Day-Ahead Market: Coverage}\label{validityDAM}
\begin{figure}[ht!]
\centering
    \begin{tikzpicture}[scale=0.77]
        \begin{axis}[
            ybar,
            bar width=0.15cm,
            ylabel={Coverage},
            xtick={-2.1,-1.1,-0.1,0.9, 2.55,3.55,4.55,5.55, 7.25,8.25,9.25,10.25,12,13,14,15,16.7,17.7,18.7,19.7,21.4,22.4,23.4,24.4,26.1,27.1,28.1, 29.1},
            xticklabels={KNN, LEAR, RF, LGBM, KNN, LEAR, RF, LGBM, KNN, LEAR, RF, LGBM, KNN, LEAR, RF, LGBM, KNN, LEAR, RF, LGBM, KNN, LEAR, RF, LGBM, KNN, LEAR, RF, LGBM},
            xticklabel style={rotate=85, anchor=east, align=center},
            legend style={at={(0.5,0.98)}, anchor=north,legend columns=-1},
            legend cell align={left},
            ymin=0.4, ymax=1, 
            width=13.5cm,
            height=6cm,
        ]

        \addplot[draw=blue!90!black!50, fill=blue!50!blue!90, fill opacity=0.6] coordinates {
                (0,0.64) (1,0.65) (2,0.66) (3,0.54)};
        \addplot[draw=blue!75!black!50, fill=blue!40!blue!75, fill opacity=0.6] coordinates {
                (4,0.74) (5,0.86) (6,0.81) (7,0.85)};
        \addplot[draw=blue!60!black!50, fill=blue!40!blue!60, fill opacity=0.6] coordinates {
                (8, 0.87) (9, 0.87) (10, 0.81) (11, 0.79)};
        \addplot[draw=blue!45!black!50, fill=blue!30!blue!45, fill opacity=0.6] coordinates {
                (12,0.85) (13,0.88) (14,0.75) (15,0.82)};                
        \addplot[draw=black!30!black!50, fill=blue!30!blue!30, fill opacity=0.6] coordinates {
                (16, 0.66) (17, 0.63) (18, 0.87) (19, 0.75)};                
        \addplot[draw=black!20!black!50, fill=blue!20!blue!20, fill opacity=0.6] coordinates {
                (20, 0.64) (21, 0.63) (22, 0.87) (23, 0.76)};                
        \addplot[draw=blue!10!black!50, fill=blue!20!blue!10, fill opacity=0.6] coordinates {
                (24,0.88) (25,0.84) (26,0.71) (27,0.78)};

        \draw[dashed, black] (axis cs:\pgfkeysvalueof{/pgfplots/xmin},0.8) -- (axis cs:\pgfkeysvalueof{/pgfplots/xmax},0.8);
        
        
    \legend{QR, SCP, EnbPI, SPCI, QRA-R, QRA-CP, Q-Ens}
    \end{axis}
    \end{tikzpicture}
    \caption{DAM Coverage for 0.1-0.9 quantile pair}
    \label{fig:QP1DAM}
\end{figure}
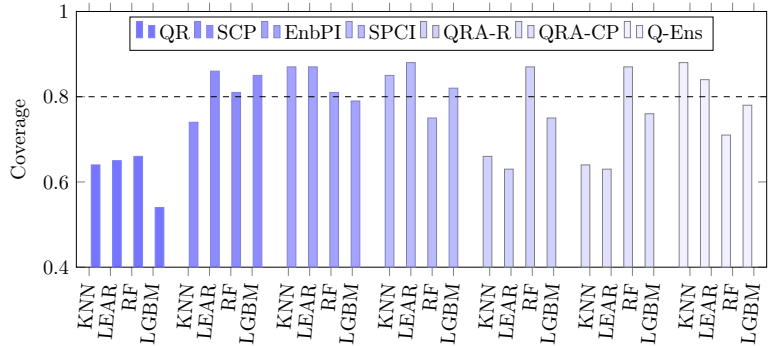

Figure \ref{fig:QP1DAM} summarizes coverage across the 0.1-0.9 quantile range in the DAM, highlighting each forecasting approaches ability to achieve coverage.
In the DAM context, aiming for a coverage of 0.8 within the 0.1-0.9 quantile range, CP models consistently outperform QR counterparts. QR produces the worst results by a margin, failing to achieve coverage in a single model. Both QRA approaches begin to diverge from QR, improving coverage for most models, LGBM in particular, and achieving the desired 0.8 coverage for RF in both QRA-R and QRA-CP.
Our Q-Ens approach yields slightly inferior results compared to EnbPI and SPCI. However, considering the lower APS for RF and LGBM models compared to EnbPI and SPCI, as shown in Table \ref{APS_DAM}, this trade-off is justified.

Focusing on the 0.3-0.7 quantile pair with a target of 0.4 in Figure \ref{fig:QP1DAM}, we see a similar trend to Figure \ref{fig:QP1DAM} with QR struggling and QRA improving QR for each model, with the exception of RF this time.
\begin{figure}[ht!]
    \centering
    \begin{tikzpicture}[scale=0.77]
        \begin{axis}[
            ybar,
            bar width=0.15cm,
            ylabel={Coverage},
            xtick={-2.1,-1.1,-0.1,0.9, 2.55,3.55,4.55,5.55, 7.25,8.25,9.25,10.25,12,13,14,15,16.7,17.7,18.7,19.7,21.4,22.4,23.4,24.4,26.1,27.1,28.1, 29.1},
            xticklabels={KNN, LEAR, RF, LGBM, KNN, LEAR, RF, LGBM, KNN, LEAR, RF, LGBM, KNN, LEAR, RF, LGBM, KNN, LEAR, RF, LGBM, KNN, LEAR, RF, LGBM, KNN, LEAR, RF, LGBM},
            xticklabel style={rotate=85, anchor=east, align=center},
            legend style={at={(0.5,0.98)}, anchor=north,legend columns=-1},
            legend cell align={left},
            ymin=0, ymax=1, 
            width=13.5cm,
            height=6cm,
            ylabel near ticks,
            yticklabel pos=left, 
        ]
    \addplot[draw=blue!95!black!50, fill=blue!95!blue!95, fill opacity=0.6] coordinates {
            (0,0.29) (1,0.23) (2,0.64) (3,0.22)};
    \addplot[draw=blue!80!black!50, fill=blue!80!blue!80, fill opacity=0.6] coordinates {
            (4,0.43) (5,0.66) (6,0.61) (7,0.64)};
    \addplot[draw=blue!65!black!50, fill=blue!65!blue!65, fill opacity=0.6] coordinates {
            (8, 0.57) (9, 0.60) (10, 0.47) (11, 0.46)};
    \addplot[draw=blue!50!black!50, fill=blue!50!blue!50, fill opacity=0.6] coordinates {
            (12,0.66) (13,0.67) (14,0.46) (15,0.56)};
    \addplot[draw=blue!40!black!50, fill=blue!40!blue!40, fill opacity=0.6] coordinates {
            (16, 0.30) (17, 0.22) (18, 0.47) (19, 0.38)};            
    \addplot[draw=blue!30!black!50, fill=blue!30!blue!30, fill opacity=0.6] coordinates {
            (20, 0.30) (21, 0.22) (22, 0.48) (23, 0.39)};                
    \addplot[draw=blue!20!black!50, fill=blue!20!blue!20, fill opacity=0.6] coordinates {
            (24,0.63) (25,0.63) (26,0.45) (27,0.51)};

        \draw[dashed, black] (axis cs:\pgfkeysvalueof{/pgfplots/xmin},0.4) -- (axis cs:\pgfkeysvalueof{/pgfplots/xmax},0.4);
        
    \legend{QR, SCP, EnbPI, SPCI, QRA-R, QRA-CP, Q-Ens}
    \end{axis}
    \end{tikzpicture}
    \caption{DAM Coverage for 0.3-0.7 quantile pair}
    \label{fig:QP2DAM}
\end{figure}
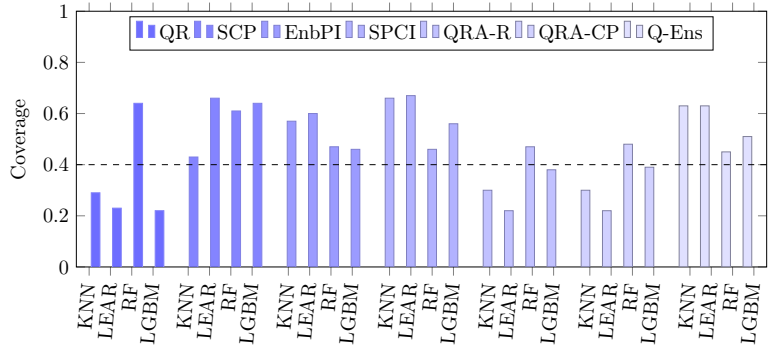
All CP approaches achieve coverage for all models this time, highlighting the advantages of CP when it comes to improving a models reliability.
Our Q-Ens achieves coverage for all models, with slightly lower scores than EnbPI and SPCI, but substantially better than QR, with the exception of RF.

\subsubsection{Efficiency \& Validity in the Day-Ahead Market: Winkler Score}\label{winklerDAM}
Table \ref{DAMWinklerScores} provides a detailed overview of Winkler Scores, highlighting each approach's ability to tie together both efficiency and validity in our chosen forecasting models.
\begin{table}[htbp]
    \centering
    \setlength{\tabcolsep}{3pt} 
    \begin{tabular}{l r r r r r r r r}
        \toprule
         Model & \multicolumn{1}{c}{QR} & \multicolumn{1}{c}{SCP} & \multicolumn{1}{c}{EnbPI} & \multicolumn{1}{c}{SPCI} & \multicolumn{1}{c}{QRA-R} &  \multicolumn{1}{c}{QRA-CP} &  \multicolumn{1}{c}{Q-Ens} \\
        \midrule
        KNN & 57.02 & \cellcolor{red!20}64.16  & 51.47 & 56.16 & 53.70 & \textbf{51.73} & 54.33 \\
        LEAR & \cellcolor{red!20}61.97 & 36.30  & 37.37 & \textbf{35.06} & 61.82 & 61.74  & 39.37 \\
        LGBM & 29.87 & 30.39  & \cellcolor{red!20}32.76 & 31.31 & 31.5 & 31.47  & \textbf{29.79} \\
        RF & \cellcolor{red!20}33.7 & 30.81  & 32.14 & 31.65 & 33.00 & 32.36  & \textbf{29.99} \\
        Avg. & 42.94 & 40.42  & 38.43 & 38.55 & \cellcolor{red!20}45.01 & 44.33  & \textbf{38.37} \\
        \bottomrule
    \end{tabular}
    \caption{Winkler Scores for the DAM. Best approach for each regressor marked in bold, worst in red.}
    \label{DAMWinklerScores}
\end{table}
Our approach Q-Ens excels producing the lowest average score, and the two lowest scores for all models with LGBM and RF.
QR and QRA pay the price of their poor performance in coverage, producing the highest average winkler scores. CP approaches produce lower winkler scores, underscoring the benefits of their strong performance in coverage, with EnbPI strong performance just missing out on our Q-Ens approach.

\subsection{Balancing Market}\label{BM_Eval}
The BM dataset include historical data and TSO predictions, revealing past market behaviors and aiding in forecasting future conditions. For the BM, we forecast prices for the next 16 open settlement periods starting at $t+2$, utilizing historical BM prices, volumes, wind discrepancies, interconnector flows, and DAM prices from the past 24-48 hours. We also incorporate future physical notifications, interconnector schedules, renewable forecasts, demand forecasts, and DAM prices for the next 8 hours. Despite similar average prices with the DAM (BM: \euro54.772/MWh, DAM: \euro53.77/MWh), the BM exhibits higher price volatility (standard deviation: \euro64.60/MWh) compared to the DAM (\euro37.38/MWh). More Details surrounding both the DAM and BM datasets can be found at \cite{OCONNOR2024101436}.

\subsubsection{Efficiency in the Balancing Market: Pinball Score \&  Interval Width}\label{sharpnessBM}
Analyzing the APS for BM models in Table \ref{BM_APS} follows a similar trend to Table \ref{APS_DAM} with Q-Ens showing strong performance producing the lowest average this time, except we see EnbPI and SPCI, two of the top approaches for the DAM, produce the two highest average scores. 
\begin{table}[ht!]
    \centering
    \setlength{\tabcolsep}{2pt} 
    \begin{tabular}{l r r r r r r r r}
        \toprule
         Model & \multicolumn{1}{c}{QR} & \multicolumn{1}{c}{SCP} & \multicolumn{1}{c}{EnbPI} & \multicolumn{1}{c}{SPCI} & \multicolumn{1}{c}{QRA-R} &  \multicolumn{1}{c}{QRA-CP} &  \multicolumn{1}{c}{Q-Ens} \\
        \midrule
        KNN  & 14.09 &  15.19 & 14.60  & \cellcolor{red!20}15.84 & 13.66  & \textbf{13.49} & 14.25 \\
        LEAR & 15.22 & \textbf{12.91}  &  13.96 & 13.79 & \cellcolor{red!20}15.33  & 15.22 &  13.21 \\
        LGBM & 12.48 & 12.87  &  13.55 & \cellcolor{red!20}13.6  & 12.58  & \textbf{12.41} &  12.70 \\
        RF   & \textbf{12.66} & 13.01  &  13.34 & \cellcolor{red!20}13.48 &  12.79 & 12.89 & 12.88 \\
        Avg. & 13.61 &  13.50 & 13.91  & \cellcolor{red!20}14.18 & 13.59  & 13.50 &  \textbf{13.26} \\
        \bottomrule
    \end{tabular}
    \caption{APS for the BM. Best approach for each regressor marked in bold, worst in red.}
    \label{BM_APS}
\end{table}
SCP produces markedly better scores, with QR and QRA approaches also performing better than EnbPI and SPCI. The two lowest scores are achieved by LGBM with QRA-R and RF with QR.

\begin{figure}[ht!]
\centering
\begin{tikzpicture}
\begin{axis}[
    samples=101,
    width=12.0cm,
    height=6.5cm,
    boxplot/draw direction=y,
    ylabel={Interval Width},
    ytick={0,10, 20, 30, 40, 50, 60, 70, 80, 90, 100}, 
    ymin=0, 
    ymax=114, 
    xtick={1,2,3,4,5,6,7,8,9,10,11,12,13,14,15,16, 17, 18, 19, 20, 21, 22, 23, 24, 25, 26, 27, 28, 29, 30, 31, 32, 33, 34},
    xticklabels={KNN, LEAR, LGBM, RF, , KNN, LEAR, LGBM, RF, , KNN, LEAR, LGBM, RF, , KNN, LEAR, LGBM, RF, ,KNN, LEAR, LGBM, RF, , KNN, LEAR, LGBM, RF, , KNN, LEAR, LGBM, RF},
    xticklabel style={rotate=85, anchor=east, align=left, font=\footnotesize}, 
    boxplot/every box/.style={
        draw=black, 
        solid  
    },
    boxplot/every median/.style={
        black, 
        thick  
    },
    xmin=0.5, 
    xmax=34.5  
]

\addplot+ [
    boxplot prepared={
        draw position=1,
        lower whisker=20.195,
        lower quartile=20.195,
        median=50,
        upper quartile=79.805,
        upper whisker=79.805,
    },
    fill=red!90
] coordinates {};

\addplot+ [
    boxplot prepared={
        draw position=2,
        lower whisker=20.395,
        lower quartile=20.395,
        median=50,
        upper quartile=79.605,
        upper whisker=79.605,
    },
    fill=red!90,
    solid  
] coordinates {};

\addplot+ [
    boxplot prepared={
        draw position=3,
        lower whisker=19.96,
        lower quartile=19.96,
        median=50,
        upper quartile=80.04,
        upper whisker=80.04,
    },
    fill=red!90,
    solid  
] coordinates {};

\addplot+ [
    boxplot prepared={
        draw position=4,
        lower whisker=13.785,
        lower quartile=13.785,
        median=50,
        upper quartile=86.215,
        upper whisker=86.215,
    },
    fill=red!90,
    solid  
] coordinates {};

\addplot+ [
    boxplot prepared={
        draw position=5,
        lower whisker=50,
        lower quartile=50,
        median=50,
        upper quartile=50,
        upper whisker=50,
    },
    fill=blue!20
] coordinates {};

\addplot+ [
    boxplot prepared={
        draw position=6,
        lower whisker=10.98,
        lower quartile=10.98,
        median=50,
        upper quartile=89.02,
        upper whisker=89.02,
    },
    fill=red!75,
] coordinates {};

\addplot+ [
    boxplot prepared={
        draw position=7,
        lower whisker=12.9,
        lower quartile=12.9,
        median=50,
        upper quartile=87.1,
        upper whisker=87.1,
    },
    fill=red!75,
    solid  
] coordinates {};

\addplot+ [
    boxplot prepared={
        draw position=8,
        lower whisker=16.17,
        lower quartile=16.17,
        median=50,
        upper quartile=83.83,
        upper whisker=83.83,
    },
    fill=red!75,
    solid  
] coordinates {};

\addplot+ [
    boxplot prepared={
        draw position=9,
        lower whisker=18.01,
        lower quartile=18.01,
        median=50,
        upper quartile=81.99,
        upper whisker=81.99,
    },
    fill=red!75,
    solid  
] coordinates {};

\addplot+ [
    boxplot prepared={
        draw position=10,
        lower whisker=50,
        lower quartile=50,
        median=50,
        upper quartile=50,
        upper whisker=50,
    },
    fill=blue!20
] coordinates {};

\addplot+ [
    boxplot prepared={
        draw position=11,
        lower whisker=1.09,
        lower quartile=1.09,
        median=50,
        upper quartile=98.91,
        upper whisker=98.91,
    },
    fill=red!60,
    solid  
] coordinates {};

\addplot+ [
    boxplot prepared={
        draw position=12,
        lower whisker=2.055,
        lower quartile=2.055,
        median=50,
        upper quartile=97.945,
        upper whisker=97.945,
    },
    fill=red!60,
    solid  
] coordinates {};

\addplot+ [
    boxplot prepared={
        draw position=13,
        lower whisker=3.305,
        lower quartile=3.305,
        median=50,
        upper quartile=96.695,
        upper whisker=96.695,
    },
    fill=red!60
] coordinates {};

\addplot+ [
    boxplot prepared={
        draw position=14,
        lower whisker=12.04,
        lower quartile=12.04,
        median=50,
        upper quartile=87.96,
        upper whisker=87.96,
    },
    fill=red!60
] coordinates {};

\addplot+ [
    boxplot prepared={
        draw position=15,
        lower whisker=50,
        lower quartile=50,
        median=50,
        upper quartile=50,
        upper whisker=50,
    },
    fill=blue!20
] coordinates {};

\addplot+ [
    boxplot prepared={
        draw position=16,
        lower whisker=12.01,
        lower quartile=12.01,
        median=50,
        upper quartile=87.99,
        upper whisker=87.99,
    },
    fill=red!45,
    solid  
] coordinates {};

\addplot+ [
    boxplot prepared={
        draw position=17,
        lower whisker=13.49,
        lower quartile=13.49,
        median=50,
        upper quartile=86.51,
        upper whisker=86.51,
    },
    fill=red!45,
    solid  
] coordinates {};

\addplot+ [
    boxplot prepared={
        draw position=18,
        lower whisker=16.87,
        lower quartile=16.87,
        median=50,
        upper quartile=83.13,
        upper whisker=83.13,
    },
    fill=red!45,
    solid  
] coordinates {};

\addplot+ [
    boxplot prepared={
        draw position=19,
        lower whisker=13.895,
        lower quartile=13.895,
        median=50,
        upper quartile=86.105,
        upper whisker=86.105,
    },
    fill=red!45,
    solid  
] coordinates {};

\addplot+ [
    boxplot prepared={
        draw position=20,
        lower whisker=50,
        lower quartile=50,
        median=50,
        upper quartile=50,
        upper whisker=50,
    },
    fill=blue!20
] coordinates {};

\addplot+ [
    boxplot prepared={
        draw position=21,
        lower whisker=19.495,
        lower quartile=19.495,
        median=50,
        upper quartile=80.505,
        upper whisker=80.505,
    },
    fill=red!30,
    solid  
] coordinates {};

\addplot+ [
    boxplot prepared={
        draw position=22,
        lower whisker=20.53,
        lower quartile=20.53,
        median=50,
        upper quartile=79.47,
        upper whisker=79.47,
    },
    fill=red!30,
    solid  
] coordinates {};

\addplot+ [
    boxplot prepared={
        draw position=23,
        lower whisker=18.81,
        lower quartile=18.81,
        median=50,
        upper quartile=81.19,
        upper whisker=81.19,
    },
    fill=red!30,
    solid  
] coordinates {};

\addplot+ [
    boxplot prepared={
        draw position=24,
        lower whisker=9,
        lower quartile=9,
        median=50,
        upper quartile=91,
        upper whisker=91,
    },
    fill=red!30,
    solid  
] coordinates {};

\addplot+ [
    boxplot prepared={
        draw position=25,
        lower whisker=50,
        lower quartile=50,
        median=50,
        upper quartile=50,
        upper whisker=50,
    },
    fill=blue!20
] coordinates {};

\addplot+ [
    boxplot prepared={
        draw position=26,
        lower whisker=19.44,
        lower quartile=19.44,
        median=50,
        upper quartile=80.56,
        upper whisker=80.56,
    },
    fill=red!20,
    solid  
] coordinates {};

\addplot+ [
    boxplot prepared={
        draw position=27,
        lower whisker=20.57,
        lower quartile=20.57,
        median=50,
        upper quartile=79.43,
        upper whisker=79.43,
    },
    fill=red!20,
    solid  
] coordinates {};

\addplot+ [
    boxplot prepared={
        draw position=28,
        lower whisker=18.205,
        lower quartile=18.205,
        median=50,
        upper quartile=81.795,
        upper whisker=81.795,
    },
    fill=red!20
] coordinates {};

\addplot+ [
    boxplot prepared={
        draw position=29,
        lower whisker=10.13,
        lower quartile=10.13,
        median=50,
        upper quartile=89.87,
        upper whisker=89.87,
    },
    fill=red!20
] coordinates {};

\addplot+ [
    boxplot prepared={
        draw position=30,
        lower whisker=50,
        lower quartile=50,
        median=50,
        upper quartile=50,
        upper whisker=50,
    },
    fill=blue!20
] coordinates {};

\addplot+ [
    boxplot prepared={
        draw position=31,
        lower whisker=11.1,
        lower quartile=11.1,
        median=50,
        upper quartile=88.9,
        upper whisker=88.9,
    },
    fill=red!10
] coordinates {};

\addplot+ [
    boxplot prepared={
        draw position=32,
        lower whisker=10.61,
        lower quartile=10.61,
        median=50,
        upper quartile=89.39,
        upper whisker=89.39,
    },
    fill=red!10,
    solid  
] coordinates {};

\addplot+ [
    boxplot prepared={
        draw position=33,
        lower whisker=13.38,
        lower quartile=13.38,
        median=50,
        upper quartile=86.62,
        upper whisker=86.62,
    },
    fill=red!10
] coordinates {};

\addplot+ [
    boxplot prepared={
        draw position=34,
        lower whisker=13.24,
        lower quartile=13.24,
        median=50,
        upper quartile=86.76,
        upper whisker=86.76,
    },
    fill=red!10,
    solid  
] coordinates {};

\node[anchor=north east, font=\footnotesize, fill=white, draw=black, text width=9.6cm, align=left] 
    at (axis cs:33.50, 114) {
        \parbox{12.0cm}{%
            \textcolor{red!90}{\rule{0.1cm}{0.2cm}} \textcolor{red!90}{\rule{0.1cm}{0.15cm}} QR 
            \textcolor{red!75}{\rule{0.1cm}{0.2cm}} \textcolor{red!75}{\rule{0.1cm}{0.15cm}} SCP
            \textcolor{red!60}{\rule{0.1cm}{0.2cm}} \textcolor{red!60}{\rule{0.1cm}{0.15cm}} EnbPI
            \textcolor{red!45}{\rule{0.1cm}{0.2cm}} \textcolor{red!45}{\rule{0.1cm}{0.15cm}} SPCI 
            \textcolor{red!30}{\rule{0.1cm}{0.2cm}} \textcolor{red!30}{\rule{0.1cm}{0.15cm}} QRA-R 
            \textcolor{red!20}{\rule{0.1cm}{0.2cm}} \textcolor{red!20}{\rule{0.1cm}{0.15cm}} QRA-CP 
            \textcolor{red!10}{\rule{0.1cm}{0.2cm}} \textcolor{red!10}{\rule{0.1cm}{0.15cm}} Q-Ens 
        }
    };

\end{axis}
\end{tikzpicture}
\caption{Interval Width for each model in the BM}
\label{IWBM}
\end{figure}
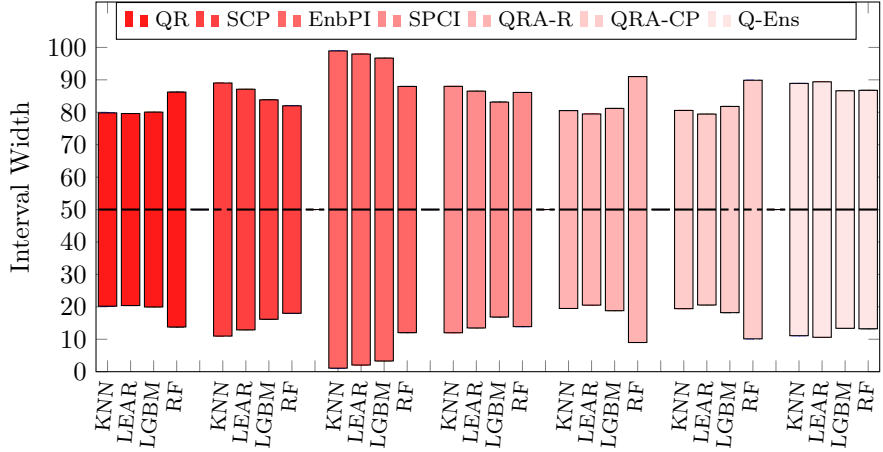

Examining efficiency in the BM, we evaluate the interval width of the models in Figure \ref{IWBM}. Compared to the average interval width for the DAM in Figure \ref{IWDAM}, all models in the BM have significantly wider interval widths, reflecting the greater inherent uncertainty in the BM.
We see the same trend as the DAM with interval width being independent of model accuracy for QR and QRA approaches, but again a strong correlation between interval width and model accuracy for all CP approaches, albeit less pronounced for SPCI than in the DAM. SPCI produces narrower interval widths than EnbPI as expected, but much more pronounced than in the DAM. Q-Ens produces wider interval widths than SPCI and QR, skewed by EnbPI large interval width.

\subsubsection{Validity in the Balancing Market: Coverage}\label{validityBM}
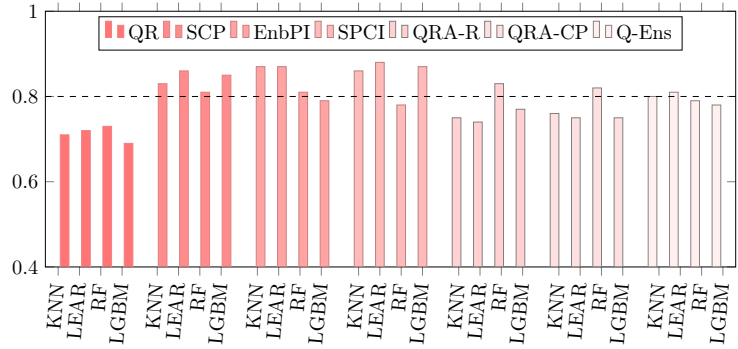
\begin{figure}[ht!]
\centering
    \begin{tikzpicture}[scale=0.77]
        \begin{axis}[
            ybar,
            bar width=0.15cm,
            ylabel={ },
            xtick={-2.1,-1.1,-0.1,0.9, 2.55,3.55,4.55,5.55, 7.25,8.25,9.25,10.25,12,13,14,15,16.7,17.7,18.7,19.7,21.4,22.4,23.4,24.4,26.1,27.1,28.1, 29.1},
            xticklabels={KNN, LEAR, RF, LGBM, KNN, LEAR, RF, LGBM, KNN, LEAR, RF, LGBM, KNN, LEAR, RF, LGBM, KNN, LEAR, RF, LGBM, KNN, LEAR, RF, LGBM, KNN, LEAR, RF, LGBM},
            xticklabel style={rotate=85, anchor=east, align=center},
            legend style={at={(0.5,0.98)}, anchor=north,legend columns=-1},
            legend cell align={left},
            ymin=0.4, ymax=1, 
            width=13.5cm,
            height=6cm,
            ylabel near ticks,
            yticklabel pos=left, 
        ]

        \addplot[draw=red!90!black!50, fill=red!50!red!90, fill opacity=0.6] coordinates {
                (0,0.71) (1,0.72) (2,0.73) (3,0.69)};
        \addplot[draw=red!75!black!50, fill=red!40!red!75, fill opacity=0.6] coordinates {
                (4,0.83) (5,0.86) (6,0.81) (7,0.85)};
        \addplot[draw=red!60!black!50, fill=red!40!red!60, fill opacity=0.6] coordinates {
                (8, 0.87) (9, 0.87) (10, 0.81) (11, 0.79)};
        \addplot[draw=red!45!black!50, fill=red!30!red!45, fill opacity=0.6] coordinates {
                (12,0.86) (13, 0.88) (14, 0.78) (15,0.87)};
        \addplot[draw=black!30!black!50, fill=red!30!red!30, fill opacity=0.6] coordinates {
                (16, 0.75) (17, 0.74) (18, 0.83) (19, 0.77)};                
        \addplot[draw=black!20!black!50, fill=red!20!red!20, fill opacity=0.6] coordinates {
                (20, 0.76) (21, 0.75) (22, 0.82) (23, 0.75)};                
        \addplot[draw=red!10!black!50, fill=red!20!red!10, fill opacity=0.6] coordinates { 
                (24,0.80) (25,0.81) (26,0.79) (27,0.78)};

        \draw[dashed, black] (axis cs:\pgfkeysvalueof{/pgfplots/xmin},0.8) -- (axis cs:\pgfkeysvalueof{/pgfplots/xmax},0.8);
        
        
    \legend{QR, SCP, EnbPI, SPCI, QRA-R, QRA-CP, Q-Ens}
    \end{axis}
    \end{tikzpicture}
    \caption{BM Coverage for 0.1-0.9 quantile pair}
    \label{fig:QP1BM}
\end{figure}

Figure \ref{fig:QP1BM} summarizes coverage across the 0.1-0.9 quantile range in the BM. Aiming for a coverage of 0.8 within this range, all models in the BM come significantly closer to achieving the desired coverage compared to their performance in the DAM.
CP approaches similarly excel, with all models achieving coverage, excluding LGBM for EnbPI and RF for SPCI, both just missing the coverage target.
QRA again improves all QR models noticeably, with RF for both QRA approaches achieving coverage. Our Q-Ens fails to perform as well as any CP approach, but outperforms all QR and QRA models bar RF for QRA.

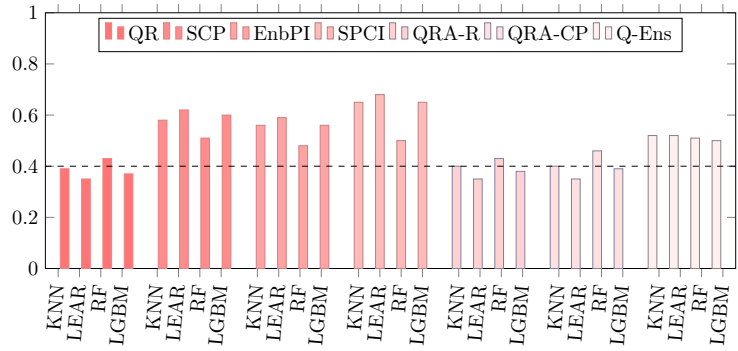
\begin{figure}[ht!]
    \centering
    \begin{tikzpicture}[scale=0.77]
        \begin{axis}[
            ybar,
            bar width=0.15cm,
            ylabel={ },
            xtick={-2.1,-1.1,-0.1,0.9, 2.55,3.55,4.55,5.55, 7.25,8.25,9.25,10.25,12,13,14,15,16.7,17.7,18.7,19.7,21.4,22.4,23.4,24.4,26.1,27.1,28.1, 29.1},
            xticklabels={KNN, LEAR, RF, LGBM, KNN, LEAR, RF, LGBM, KNN, LEAR, RF, LGBM, KNN, LEAR, RF, LGBM, KNN, LEAR, RF, LGBM, KNN, LEAR, RF, LGBM, KNN, LEAR, RF, LGBM},
            xticklabel style={rotate=85, anchor=east, align=center},
            legend style={at={(0.5,0.98)}, anchor=north,legend columns=-1},
            legend cell align={left},
            ymin=0, ymax=1, 
            width=13.5cm,
            height=6cm,
            ylabel near ticks,
            yticklabel pos=left, 
        ]
        \addplot[draw=red!90!black!50, fill=red!50!red!90, fill opacity=0.6] coordinates {
                (0,0.39) (1,0.35) (2,0.43) (3,0.37)};
        \addplot[draw=red!75!black!50, fill=red!40!red!75, fill opacity=0.6] coordinates {
                (4,0.58) (5,0.62) (6,0.51) (7,0.6)};
        \addplot[draw=red!60!black!50, fill=red!40!red!60, fill opacity=0.6] coordinates {
                (8, 0.56) (9, 0.59) (10, 0.48) (11, 0.56)};
        \addplot[draw=red!45!black!50, fill=red!30!red!45, fill opacity=0.6] coordinates {
                (12,0.65) (13,0.68) (14,0.5) (15,0.65)};                
        \addplot[draw=blue!30!black!50, fill=red!30!red!30, fill opacity=0.6] coordinates {
                (16, 0.40) (17, 0.35) (18, 0.43) (19, 0.38)};                
        \addplot[draw=blue!20!black!50, fill=red!20!red!20, fill opacity=0.6] coordinates {
                (20, 0.40) (21, 0.35) (22, 0.46) (23, 0.39)};                
        \addplot[draw=red!10!black!50, fill=red!10!red!10, fill opacity=0.6] coordinates { 
                (24,0.52) (25,0.52) (26,0.51) (27,0.5)};
        
        \draw[dashed, black] (axis cs:\pgfkeysvalueof{/pgfplots/xmin},0.4) -- (axis cs:\pgfkeysvalueof{/pgfplots/xmax},0.4);
        
    \legend{QR, SCP, EnbPI, SPCI, QRA-R, QRA-CP, Q-Ens}
    \end{axis}
    \end{tikzpicture}
    \caption{BM Coverage for 0.3-0.7 quantile pair}
    \label{fig:QP2BM}
\end{figure}

Focusing on the 0.3-0.7 quantile pair aiming for a coverage of 0.4 in Figure \ref{fig:QP2BM}, again all CP approaches achieve coverage, including Q-Ens. 
QRA fails to demonstrate a noticeable improvement off of QR, but all models come close to achieving coverage, with RF for QR and both QRA achieving the 0.4 target.

\subsubsection{Efficiency \& Validity in the Balancing Market: Winkler Score}\label{winklerBM}
Table \ref{BMWinklerScores} provides a overview for Winkler Scores in the BM, offering insights into each approaches ability to tie efficiency and validity together.
\begin{table}[htbp]
    \centering
    \setlength{\tabcolsep}{3pt} 
    \begin{tabular}{l r r r r r r r r}
        \toprule
         Model & \multicolumn{1}{c}{QR} & \multicolumn{1}{c}{SCP} & \multicolumn{1}{c}{EnbPI} & \multicolumn{1}{c}{SPCI} & \multicolumn{1}{c}{QRA-R} &  \multicolumn{1}{c}{QRA-CP} &  \multicolumn{1}{c}{Q-Ens} \\
        \midrule
        \midrule
        KNN  & 124.18 & \cellcolor{red!20}134.77  & 131.35 & 132.18 & \textbf{119.21} & 120.89  & 126.35 \\
        LEAR & 127.79 & \textbf{115.02}  & 122.63 & 126.16 & \cellcolor{red!20}129.03 & 127.75  & 124.09 \\
        LGBM & \textbf{107.34} & 115.77  & 111.18 & \cellcolor{red!20}122.41 & 109.47 & 108.34  & 111.18 \\
        RF   & 114.37 & 113.88  & 111.84 & 114.32 & \cellcolor{red!20}116.15 & 115.07  & \textbf{111.39}\\
        Avg. & 118.42 & 119.86  & 119.25 & \cellcolor{red!20}123.77 & 118.47 & \textbf{118.01}  & 118.25 \\
        \bottomrule
    \end{tabular}
    \caption{Winkler Scores for the BM. Best approach for each regressor marked in bold, worst in red.}
    \label{BMWinklerScores}
\end{table}
The analysis in Table \ref{BMWinklerScores} highlights the strong performance of our ensemble approach again, resulting in the second lowest winkler score, slightly underperforming QRA-CP. QR and QRA demonstrate considerably improved performance than they did for winkler scores in the DAM, with QRA-CP coming first, QR third, and QRA-R fourth. The last 3 spots are occupied by all three CP approaches with SPCI changing from just 0.07 off the top approach in the DAM, to the worst performer in the BM. QR achieves the lowest score of any model with LGBM, while Q-Ens has the second lowest with RF. These results are likely due to the closer coverage scores and narrower interval widths for the QR and QRA approaches.

\subsection{Financial Performance Analysis}\label{financialperf}
For our financial evaluation, we utilise a Battery Energy Storage System (BESS). This section outlines our BESS trading strategy, crucial for ensuring grid stability and seamlessly integrating renewable energy into the dynamic energy landscape. This consists of our Single trade strategy (TS1) and our high-frequency trading strategy (TS2):
\begin{itemize}
    \item TS1 is a rule-based heuristic trading strategy adopted from \cite{uniejewski2021regularized, uniejewski2023smoothing}. This strategy utilizes quantile-based forecasts to optimize trading decisions involving a hypothetical 1 MWh battery with no discharge limit, 80\% discharge efficiency, and 98\% charge efficiency. Over the specified time horizon (1 day for DAM or 8 hours for BM), a single buy-sell pair trade is permitted, with the requirement that the buy trade occurs before the sell trade.

    \item TS2 leverages optimization techniques to maximize profits in electricity markets, adopted from \cite{oconnor2024optimizing}. It incorporates flexible buy/sell timestamps within bottleneck-constrained volumes, considering realistic constraints such as battery capacity, charging speed, discharging rate, minimum charge level, and energy efficiency. The dynamic adjustment of charging and discharging operations, based on forecasted minimum and maximum price periods, is a fundamental aspect of this adaptive approach. The overarching goal of TS2 is to enhance profitability within bottleneck limitations, significantly increasing trading frequency and contributing to overall market efficiency.
\end{itemize}
For more detailed information on the implementation of both TS1 and TS2 see  \cite{oconnor2024optimizing}.
\begin{table}[ht!]
    \centering
    \setlength{\tabcolsep}{0.9pt} 
    \begin{tabular}{l|*{6}{>{\centering\arraybackslash}p{1.68cm}}}
        \toprule
        \textbf{Model} & \textbf{TS1 DAM} & \textbf{TS1 BM} & \textbf{TS2 DAM} & \textbf{TS2 BM} & \textbf{Avg. DAM} & \textbf{Avg. BM} \\
        \midrule
        QR    & \euro16,376                    & \euro3,634    
              &                   \euro21,887  & \euro16,112    
              &                   \euro19,131                    & \euro9,873   \\

        SCP    & \euro16,032 & \cellcolor{red!20}-\euro1,215 
               & \cellcolor{red!20}\euro17,610 & \cellcolor{red!20}\euro11,119  
               &  \cellcolor{red!20}\euro16,821 & \cellcolor{red!20}\euro4,952 \\
        
        EnbPI             & \euro16,806  & \euro3,214  
        & \euro20,932  &                   \euro16,237    
        & \euro18,869  & \euro9,725   \\
        
        SPCI    & \euro16,486  & \euro3,280  
                & \euro21,632  & \euro16,807    
                & \euro19,059  & \euro10,044   \\

        QRA-R    & \cellcolor{red!20}\euro13,868  & \euro2,768
                 & \euro20,938  & \euro14,705 
                 & \euro17,403  & \euro8,736     \\

        QRA-CP   & \euro13,880  & \textbf{\euro3,923}
                 & \euro20,805  & \euro15,061 
                 & \euro17,342  & \euro9,492   \\

        Q-Ens    & \textbf{\euro16,935}  & \euro3,748 
                 & \textbf{\euro21,926}  & \textbf{\euro17,400}    
                 & \textbf{\euro19,430}  & \textbf{\euro10,574}   \\
        \bottomrule
    \end{tabular}
    \caption{Financial Performance Comparison of RF \& LGBM Models. Best approach marked in bold, worst in red.}
    \label{TS1TS2Results}
\end{table}
Table \ref{TS1TS2Results} summarizes the financial performance of various forecasting approaches in the DAM and BM for trading strategies TS1 and TS2, excluding the KNN and LEAR models to focus on top performers. SCP consistently produces the worst results, with the lowest averages and the poorest performance in TS2 for both markets and TS1 for the BM. In contrast, EnbPI and SPCI demonstrate strong performance, with SPCI leading among the CP approaches, highlighting the benefits of time series-adapted methods. QRA does not surpass QR, with QR achieving better average scores in both markets and QRA-R ranking last for TS1 in the DAM. However, QRA-CP secures the top result for TS1 in the BM and scores closer to QR in the BM. The most notable outcome is the strong performance of our ensemble approach, which achieves the highest average result in both the DAM and BM, only missing the top position for TS1 in the BM.

\subsection{Key Findings and Limitations}
The application of CP techniques, particularly EnbPI and SPCI, improves the accuracy of PEPF in both DAM and BM, outperforming traditional QR models by providing more reliable PIs with consistent coverage. The ensemble approach, combining QR with EnbPI and SPCI, improves the balance between narrow intervals and coverage, resulting in more consistent and accurate forecasts. Financial analysis of these forecasting methods indicates that CP techniques, particularly our ensemble, Q-Ens, can increase profitability in trading algorithms; however, their dependence on large amounts of historical data and higher computational requirements may limit their practicality in real-time scenarios. These models can also face challenges when encountering extreme price events or sudden market shifts.



\section{Conclusion}\label{conclusionsec}
This study evaluated several probabilistic forecasting methods for electricity prices in the Irish DAM and BM, focusing on QR, QRA-R, QRA-CP, SCP, EnbPI, SPCI, and the proposed ensemble method Q-Ens. Our evaluation covered multiple dimensions—accuracy, reliability, interval width, coverage, and financial performance. We found that QR and QRA approaches performed well in efficiency metrics but fell short in coverage, where CP methods like SCP, EnbPI, and SPCI excelled. The ensemble approach (Q-Ens) demonstrated strong results in both efficiency and validity, and it delivered superior financial outcomes in a single trade and high-frequency trading strategy, confirming its practical value for market participants.
The integration of CP techniques into EPF offers a promising way to improve forecast reliability and financial outcomes, especially as the energy market evolves and renewable energy integration increases. However, the financial evaluation, based on simulated trading, may not fully capture the complexities of real-world market conditions, including transaction costs, market liquidity, and sudden structural shifts. Therefore, further research should focus on testing these methods in diverse markets, including IDM, and exploring the practical limitations of CP models, particularly in environments with extreme price events or limited historical data.
Future studies could also delve deeper into more granular 5-minute BM datasets and incorporate probabilistic forecasting frameworks that address transaction costs and market conditions. Additionally, improving the robustness of CP methods, particularly with respect to outliers and non-exchangeable time series, will be fundamental. Exploring the application of CP models in renewable energy forecasting, smart grid management, and real-time data streams presents a rich avenue for future development, which will help to create more adaptable and reliable forecasting systems in increasingly complex energy markets.

\section*{Acknowledgments}
This work was conducted with the financial support of Science Foundation Ireland under Grant Nos. 18/CRT/6223 and 12/RC/2289-P2 which are co-funded under the European Regional Development Fund. For the purpose of Open Access, the author has applied a CC BY public copyright licence to any Author Accepted Manuscript version arising from this submission.
\bibliography{mybibfile}

\section*{Declaration of generative AI and AI-assisted technologies in the writing process}

During the preparation of this work, the authors used ChatGPT to improve the paper's readability. After using this tool/service, the authors reviewed and edited the content as needed and take full responsibility for the content of the publication.

\end{document}